\documentclass{article}

\PassOptionsToPackage{square,sort,comma,numbers}{natbib}

 \usepackage[preprint]{neurips_2020}

\usepackage[utf8]{inputenc}
\usepackage[T1]{fontenc}
\usepackage{hyperref}       
\usepackage{url}            
\usepackage{booktabs}       
\usepackage{amsfonts}       
\usepackage{nicefrac}       
\usepackage{microtype}      
\usepackage{graphicx}
\usepackage{subfigure}
\usepackage{amsmath}
\usepackage{makecell}
\usepackage{multirow}
\usepackage{amsfonts}
\usepackage{varwidth}
\usepackage{caption}
\usepackage{amsthm,amsmath,amssymb}
\usepackage{mathrsfs}
\usepackage{bm}
\usepackage{color,xcolor}
\usepackage{booktabs} 

\title{Rotation-Equivariant Neural Networks for Privacy Protection}

\author{%
Hao Zhang\\
Shanghai Jiao Tong University\\
\texttt{1603023-zh@sjtu.edu.cn} \\
\And
Yiting Chen \\
Shanghai Jiao Tong University\\
\texttt{sjtucyt@sjtu.edu.cn} \\
\And
Haotian Ma \\
Southern University of Science and Technology\\
\texttt{11612807@mail.sustc.edu.cn} \\
\And
Xu Cheng \\
Shanghai Jiao Tong University\\
\texttt{xcheng8@sjtu.edu.cn} \\
\And
Qihan Ren \\
Shanghai Jiao Tong University\\
\texttt{renqihan@sjtu.edu.cn} \\
\And
Liyao Xiang \\
Shanghai Jiao Tong University\\
\texttt{xiangliyao08@sjtu.edu.cn} \\
\And
Jie Shi \\
Huawei International\\
\texttt{shi.jie1@huawei.com} \\
\And
Quanshi Zhang \\
Shanghai Jiao Tong University\\
\texttt{zqs1022@sjtu.edu.cn} \\
}
\newcommand{\tabincell}[2]{\begin{tabular}{@{}#1@{}}#2\end{tabular}}
\begin{document}

\maketitle

\begin{abstract}
In order to prevent leaking input information from intermediate-layer features, this paper proposes a method to revise the traditional neural network into the rotation-equivariant neural network (RENN). Compared to the traditional neural network, the RENN uses $d$-ary vectors/tensors as features, in which each element is a  $d$-ary number. These $d$-ary features can be rotated (analogous to the rotation of a $d$-dimensional vector) with a random angle as the encryption process. Input information is hidden in this target phase of $d$-ary features for attribute obfuscation. Even if attackers have obtained network parameters and intermediate-layer features, they cannot extract input information without knowing the target phase. Hence, the input privacy can be effectively protected by the RENN. Besides, the output accuracy of RENNs only degrades mildly compared to traditional neural networks, and the computational cost is significantly less than the homomorphic encryption.
\end{abstract}

\section{Introduction}
Considering computational constraints, raw data collected for deep learning is often processed in a distributed system rather than being processed locally. Hence, the privacy leakage happens frequently and has received much attention recently. Many researches  \cite{dosovitskiy2016inverting, zeiler2014visualizing, mahendran2015understanding, shokri2017membership,Property2018Karan, melis2018exploiting, Yang2019Adversarial} have pointed out that, attackers can recover significant amount of sensitive input information from intermediate-layer features of a DNN. As countermeasures, several studies \cite{hybirddeeplearning, PrivyNet, notjustprivacy, PrivacyZhang, SecureML} have been proposed, but have limited applicability mostly due to the exorbitant computational cost.

The task of attribute obfuscation and privacy protection on DNNs mainly needs to satisfy two requirements. First, even if attackers have access to network parameters and intermediate-layer features, they cannot reconstruct the input or infer private attributes of inputs. Second, the privacy-protection method should not increase the computational cost, or affect the task accuracy significantly.

Therefore, we propose a set of generic rules to revise a traditional neural network into a rotation-equivariant neural network (RENN) without sacrificing either task accuracy or computational efficiency too much. Unlike traditional neural networks, the RENN uses $d$-ary vectors/tensors as features, where each element is a $d$-ary number. Input information is placed in a random component of the $d$-ary feature, where other $(d-1)$ components are used as $(d-1)$ fooling counterparts.

The basic idea for privacy protection is to rotate the $d$-ary intermediate-layer feature with the same certain angle as the encryption process. In the $d$-ary feature, each element is rotated with a specific rotation angle, which is termed the \emph{phase}. This phase is analogous to the orientation of a $d$-dimensional vector. In this way, we consider that input information is hidden inside the target phase, which can be taken as the private key. Without knowing this target phase, the input information cannot be recovered from the encrypted $d$-ary feature.

\begin{figure}[t]
\begin{center}\centerline{\includegraphics[width=0.85\columnwidth]{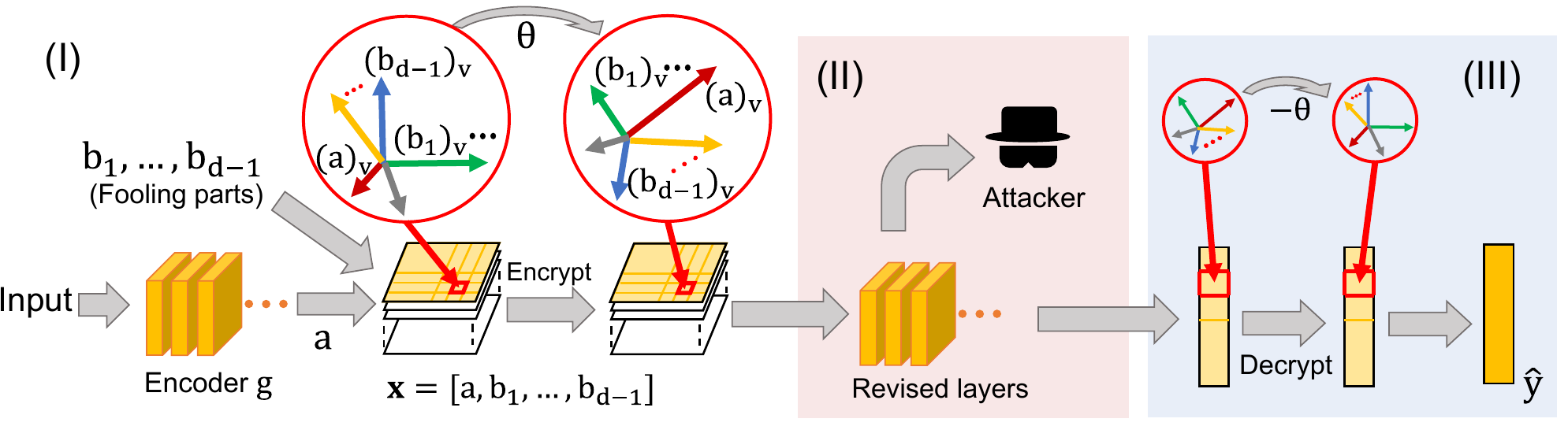}}
\caption{Overview of the RENN. RENN is composed of an encoder, a processing module, and a decoder. The encoder module (I) is located at the local device. The input feature $a$ is extracted and converted into a $d$-ary feature for encryption. The encryption process is to rotate this $d$-ary feature by a certain rotation matrix $\bm{R}$ along a random axis with a random angle $\theta$. This encrypted $d$-ary feature is further processed by the processing module (II) which satisfies the rotation equivariance property. The decoder module (III) uses the same rotation to decrypt features to obtain the final result.}
\label{fig:overview}\end{center}\vspace{-20pt}\end{figure}

The architecture of the RENN is shown in Fig.~\ref{fig:overview}, which can be split into an encoder, a processing module, and a decoder. The encoder extracts the feature of the input, which is a real-valued vector/tensor, and converts it into a certain component of the $d$-ary vector/tensor. Then, the encoder rotates this $d$-ary feature to make the input information hidden in a target phase as the encryption process. The rotated $d$-ary feature is sent to the processing module for further process. The decoder decrypts the processed feature and obtains the final result.

In order to ensure the successful decryption of the decoder, the processing module has to satisfy the rotation equivariance property \cite{cohen2016steerable}. In other words, given an input, we first rotate the feature  extracted from the input with a certain angle $\theta$, and send this rotated feature to the processing module to obtain the intermediate-layer feature $\alpha$. Alternatively, we can first send the feature extracted from the input to the processing module, and then rotate this processing module's feature with the same rotation angle $\theta$ to get the rotated feature $\beta$. The rotation equivariance property indicates that the first rotated and then processed feature $\alpha$ equals to the first processed and then rotated feature $\beta$. Meanwhile, this property also ensures that we can directly invert the rotation for encryption to decrypt the output feature of the processing module.

 In this way, we propose a set of rules to revise traditional layerwise operations including ReLU, batch-normalization, \emph{etc.} to make them rotation-equivariant. The rotation equivariance property ensures that the decoder can successfully decrypt input information. The proposed rules can be broadly used to revise layerwise operations in DNNs with different architectures for various tasks. Furthermore, in order to improve the ability of privacy protection, the target phase is obfuscated by adversarial learning, \emph{i.e.} using a GAN to generate $(d-1)$ fooling counterparts in $d$-ary features to fool the attacker. Experimental results showed that RENNs outperformed other baselines in terms of privacy protection, yet the accuracy was not significantly affected.

Previous methods usually sacrificed the computational efficiency or decrease the task accuracy for privacy protection. Siamese fine-tuning~\cite{hybirddeeplearning} reduced the level of sensitive information in the input, so that attackers could not infer private properties from intermediate-layer features. The PrivyNet~\cite{PrivyNet} was proposed  to explore the trade-off between the privacy protection and the task accuracy. A lightweight privacy protection mechanism~\cite{notjustprivacy} was applied, which consisted of data nullification and random noise addition. Homomorphic encryption is a cryptographic technique, which can be applied in the privacy protection in deep learning. The BGV encryption scheme~\cite{PrivacyZhang} was adopted to encrypt data, and the high-order backpropagation algorithm was performed for training. Data was distributed among two non-colluding servers in \cite{SecureML}, where the model was trained using secure two-party computation. Complex-valued NN~\cite{deepPrivacy} used complex-valued feature for privacy protection. In comparison, our algorithm can be considered as a generic method to revise traditional DNNs into privacy-preserving RENNs with broad applicability. Crucially, according to Table~\ref{tab:k-anonymity}, the computational cost of the homomorphic encryption is roughly $809-3236$ times than that of the RENN. Besides, the complex-valued NN can be regraded as a special case of the RENN ($d=2$), which is discussed in Sec.~\ref{2.3}.

Contributions of this study are summarized as follows. (1) We propose a set of generic rules to revise traditional DNNs into RENNs,  which hide sensitive input information in a random phase. Without knowing the target phase, attackers can hardly infer any input information from features. (2) RENNs incur far less computational overhead than crypto-based methods.

A previous and specific version of the RENN is the quaternion neural network~\cite{QNN}.

\section{Algorithm}
\subsection{Deep Rotation-Equivariant Features }
\label{3.1}
\textbf{Rotation:} For a traditional DNN, an intermediate-layer feature can be a real-valued vector/matrix/tensor. For simplicity, this real-valued feature can be vectorized into $f \in \mathbb{R}^{n}$. In this paper, we extend the real-valued feature into a $d$-ary feature, where each element is a $d$-ary number. This $d$-ary feature can be denoted by $\bm{f} \in \mathbb{H}_{d}^{n}$, where $\mathbb{H}_{d}$ represents the domain of $d$-ary numbers and $\bm{f}$ is a vector of $d$-ary numbers. In real applications, we can still apply traditional convolution operations to the $d$-ary feature, \emph{i.e.} applying convolution operations to the real-valued feature component corresponding to each component of this $d$-ary feature independently.

In this way, the rotation of the $d$-ary feature $\bm{f}$ can be performed with a rotation matrix $\bm{R} \in \mathbb{R}^{d \times d}$~\cite{hanson19954}, \emph{i.e.} $\bm{R} \circ \bm{f}$, where $\circ$ denotes the matrix multiplication between $\bm{R}$ and each $d$-ary element in $\bm{f}$. In this way, we can consider each $d$-ary element in the $d$-ary feature as the $d$-dimensional vector, and  $\bm{R} \circ \bm{f}$ indicates that we apply the same rotation matrix to each $d$-dimensional vector. $\bm{R}$ is a rotation matrix if and only if it satisfies (1) $\bm{R}$ is a unit orthogonal matrix; (2) $det(\bm{R}) =1$.

\textbf{Rotation Equvariance}: In this study, we perform layerwise revisions to a DNN to achieve rotation equvariance property for the $d$-ary feature during the forward propagation. Given an input, we first rotate the feature  extracted from the input with a certain angle, and send this rotated feature to a DNN to obtain the intermediate-layer feature $\alpha$. Meanwhile, we can first send the feature extracted from the input to the DNN, and then rotate this intermediate-layer feature with the same rotation angle to get the rotated feature $\beta$. The rotation equivariance property indicates that the first rotated and then processed feature $\alpha$ equals to the first processed and then rotated feature $\beta$. To achieve the above rotation equivariance, we should ensure the layerwise rotation equivariance. Let $\Phi(\bm{f})=\Phi_L(\Phi_{L-1}(\cdots\Phi_1(\bm{f}))$ represent cascaded layers of a DNN, where $\Phi_l(\cdot)$ and $\bm{f}_l=\Phi_l(\Phi_{l-1}(\cdots\Phi_1(\bm{f}))$ denote the function of the $l$-th layer and its output, respectively. We revise traditional layerwise operations to ensure the feature transformation in $\Phi$ rotation-equivariant as follows.
\begin{small}\begin{equation}\label{1}
\forall \;l, \;\Phi_{l}(\bm{R} \circ\bm{f}_{l-1})=\bm{R} \circ\Phi_{l}(\bm{f}_{l-1}) \qquad \Longrightarrow \qquad  \Phi(\bm{R} \circ\bm{f})=\bm{R} \circ\Phi(\bm{f})
\end{equation}\end{small}
In this way, six most widely used operations including convolution, ReLU, batch-normalization, avg/max-pooling, dropout, and skip-connection are revised to make them rotation-equivariant.

$\bullet\quad$\emph{Convolution:} For the convolutional layer, we remove the bias term to ensure rotation equivariance, and get $\text{Conv}(\bm{f})=w\otimes \bm{f}$. Note that this revision can also be applied to the fully-connected layer, since the fully-connected layer can be considered as a special convolutional layer.

$\bullet\quad$\emph{ReLU:} The ReLU operation is revised into $\text{ReLU}(\bm{f}_v)=\frac{\|\bm{f}_v\|}{\text{max}\{\|\bm{f}_v\|, C\}}\cdot\bm{f}_v$, where $C$ denotes a positive scalar, and $\bm{f}_v\in \mathbb{H}_{d}$ is the $v$-th ($1\leq v \leq n$) $d$-ary element in the feature $\bm{f} \in \mathbb{H}_{d}^{n} $.

$\bullet\quad$\emph{Batch-normalization:} The batch-normalization operation is transformed into $\text{norm}(\bm{f}_v^{(k)})=\small{\bm{f}_v^{(k)} / {\sqrt{\mathbb{E}_{k'}[\|\bm{f}_v^{(k')}\|^2]+\epsilon}}}$, where $\bm{f}_v^{(k)} \in \mathbb{H}_{d}$ represents the $v$-th element of the $k$-th sample in the batch, and $\epsilon$ denotes a small positive scalar to prevent $\bm{f}_v^{(k)}$ from being divided by zero.

$\bullet\quad$\emph{Avg/Max-pooling:} The Avg-pooling operation satisfies Eq.~\eqref{1} without any additional revisions. Whereas, the max-pooling operation selects the element of the $d$-ary feature with the largest norm value from the receptive field, which is revised into $\text{maxpool}(\bm{f})=\bm{f}\circ \bm{m}$, where $\bm{f}\in \mathbb{H}_{d}^{n}, \bm{m} \in \{0, 1\}^{n}$. If the $\hat{v}$-th element in $\bm{f}$ is selected, then $\hat{v} = \mathop{\arg\max}_{v \in \textrm{receptive}} {\|\bm{f_v}\|}_2$, and $\bm{m}_v= \bm{1} \in \mathbb{H}_{d}$; otherwise, $\bm{m}_v= \bm{0}$.

$\bullet\quad$\emph{Dropout:} The $d$-ary element $\bm{f}_v\in \mathbb{H}_{d}$ in $\bm{f}$ are randomly dropped out, \emph{i.e.} being set to a $d$-ary number of zero with a certain dropout rate.

$\bullet\quad$\emph{Skip connection:} The skip connection can be formulated as $\bm{f}+\Phi(\bm{f})$ to satisfy Eq.~\eqref{1}.\\
Above operations satisfy rotation equvariance property, please see supplementary materials for proofs.

\subsection{RENN}
\label{3.2}

\begin{figure}[t]
\begin{center}\centerline{\includegraphics[width=0.9\columnwidth]{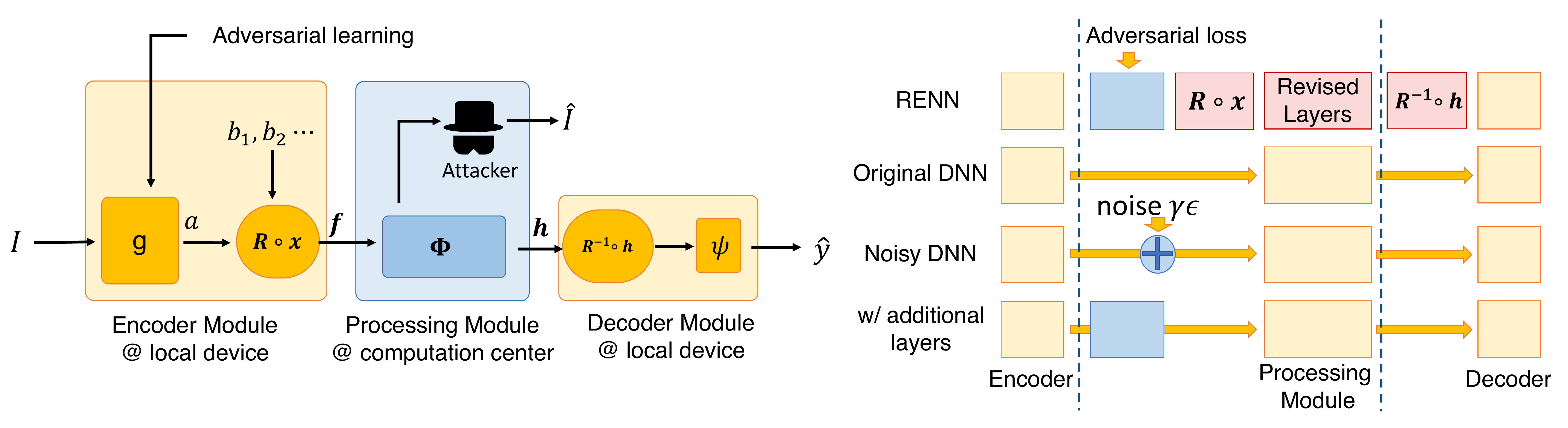}}
\caption{Architecture of the RENN (left) and alignment of architectures between neural networks (right). The RENN is divided into three modules: the encoder module, the processing module, and the decoder module. The encoder module extracts features from the input, and transforms it into a $d$-ary feature. The input information is hidden into a random phase of the $d$-ary feature. The processing module processes encrypted features without knowing the target phase. The decoder module decrypts the input information with the target phase to get the final result.}
\label{fig:structure}\end{center}\vskip -0.2in\vspace{-15pt}\end{figure}

The RENN aims to achieve privacy protection. Even if given network parameters and intermediate-layer features, attackers cannot recover the input. As Fig.~\ref{fig:structure} (left) shows, the RENN is split into three modules: (1) The \textbf{encoder module} is usually embedded inside a local device at the user end, which is used to encrypt  features extracted from the input and sends them to the processing module. (2) The \textbf{processing module} is deployed at the public cloud and aims to further process encrypted features. (3) The \textbf{decoder module} resides at the local device of the user, whose function is to receive and decrypt features from the processing module.

$\bullet\quad$\textbf{Encoder:} Given an input $I\in \textbf{I}$, as shown in Fig.~\ref{fig:overview}, the encoder module first extracts the feature $a = g(I) \in \mathbb{R}^n$ from the input, where $g$ can be implemented as a multi-layer network. Then, this input feature $a$ together with $(d-1)$ fooling counterparts are converted into a $d$-ary feature $\bm{x} \in \mathbb{H}_{d}^{n}$. These $(d-1)$ fooling counterparts, $b_1,\cdots,b_{d-1} \in \mathbb{R}^n$, are prepared by the encoder as well. Without loss of generality, the input feature $a$ is set as the first component of the $d$-ary feature, \emph{i.e.} $\bm{x} =[a, b_1, \cdots, b_{d-1}]\in \mathbb{H}_{d}^{n}$. The encrypted feature $\bm{f}$ is obtained by rotating $\bm{x}$ with a rotation matrix $\bm{R}$, \emph{i.e.} $\bm{f} = \bm{R}\circ\bm{x}$, where $\circ$ denotes the matrix multiplication between $\bm{R}$ and each $d$-ary element in $\bm{x}$. Note that, the rotation of the $d$-ary feature $\bm{x}$ is analogous to rotating the $d$-dimensional vector with the matrix $\bm{R}$. Moreover, the only phase $\theta = \bm{R}\cdot \rho$ contains the input information, so $\theta$
can be regarded as the private key, where $\rho=[1,0,0,\cdots] ^T\in \mathbb{R}^d$ is a $d$-dimensional vector.

$\bullet\quad$\textbf{Processing module:} The processing module $\Phi$ is constructed with the revised layerwise operations in Sec.~\ref{3.1}, which satisfy the rotation-equivariance property. Given the encrypted features $\bm{f}$, the processing module outputs a $d$-ary feature $\bm{h}= \Phi(\bm{f})=\Phi(\bm{R\circ x})$. Note that the processing module has no access to the target phase. Even if attackers obtain intermediate-layer features, they cannot recover the input.

$\bullet\quad$\textbf{Decoder:} Considering the rotation equivariance, the decoder $\Psi$ is able to decode intermediate-layer features $\bm{h}$ with the same rotation matrix $\bm{R}$. Either a shallow network or a simple softmax layer can be taken as the decoder, and the decryption result is  $\hat{y}=\Psi(\Gamma (\bm{R}^{-1}\circ \bm{h}))$, where $\Gamma$ denotes the operation of extracting the first component from the $d$-ary feature and returns a real-valued feature.

$\bullet\quad$\textbf{Encoder based on GAN:} To further boost the robustness to attacks, we adopt the adversarial learning and use a GAN~\cite{GAN} to train the encoder. The generator of the GAN can be implemented as the encoder module, while the discriminator is regarded as the attacker. $d$-ary features produced by the generator satisfy (1) that features contain enough information for the task; (2) that features in the target phase and features in the other phase follow the same distribution to make the discriminator hard to distinguish which phase contains the input information. The discriminator aims to learn to discern the target phase which encodes input information.

Given an input $I\in\textbf{I}$ and its real-valued feature $a=g(I) \in \mathbb{R}^n$, we concatenate $a$ and $(d-1)$ fooling counterparts to generate the encrypted feature $ \bm{f} = \bm{R\circ x}$, $\bm{x} = [a, b_1, \cdots, b_{d-1}] \in \mathbb{H}_{d}^{n}$. Theoretically, compared to $d$-ary features in an intermediate layer of the processing module, $\bm{f}$ is more close to the input feature $a$ and is the easiest to attack. Thus, we use $\bm{f}$ to train the GAN. The attacker aims to estimate the most probable phase to decrypt $\bm{f}$ and get the input feature $a$. Let $\bm{R}'$ denote the rotation estimated by the attacker, and the feature decrypted by the attacker is $ a' = \Gamma(\bm{R}'^{-1}\circ\bm{f})$. Both $a$ and $a'$ are inputs of the discriminator $D$, which needs to learn to seperate $a$ and $a'$. We adopt the WGAN~\cite{WGAN} to train the generator and the discriminator jointly as follows.
\begin{equation}\label{eqn:wgan}
\quad\min_g \max_D L(g, D)=\mathbb{E}_{I}[D(a)-\mathbb{E}_{\bm{R}'\neq\bm{R}}[D(a')]]
\end{equation}
Hence, the overall loss to optimize the RENN consists of both the GAN loss and the task loss.
\begin{equation}\label{eqn:overall_loss}
\min_{g,\Phi,\Psi}\max_{D} Loss=\min_{g, \Phi, \Psi}[\max_{D}L(g, D)+L_{\text{task}}(\hat{y}, y)]
\end{equation}
where $L_{\text{task}}$ is the loss for the task, and $y$ is the ground-truth label.

\subsection{Previous studies as special cases of RENNs}
\label{2.3}
$\bullet\quad$\emph{Specialization of RENNs}: Theoretically, the complex-valued neural network(NN)~\cite{deepPrivacy} can be taken as a special case of the RENN ($d=2$). The complex-valued NN used a complex-valued feature $\bm{x'} = a+b\bm{i}$, where $b$ is a fooling counterpart excluding input information. The encrypted feature $\bm{f'}$ is obtained by rotating $\bm{x'}$ with a random phase $\theta$, \emph{i.e.} $\bm{f'} = \textrm{exp}(\bm{i}\theta)[a+b\bm{i}]$.

In this study, we can also propose a quaternion NN (QNN), which uses quaternion-valued features for privacy protection. The QNN is also a special case of the RENN ($d=3$). The quaternion $\bm{q}=q_0+q_1 \bm{i}+q_2\bm{j}+q_3\bm{k}$ is defined in~\cite{quaternion}, which consists of three imaginary parts $q_1\bm{i}$, $q_2\bm{j}$, $q_3\bm{k}$, and one real part $q_0$. A pure quaternion has a zero value of the real part, $q_0=0$. The products of basis elements $\bm{i}$, $\bm{j}$, $\bm{k}$ satisfy $\bm{i}^2 = \bm{j}^2 = \bm{k}^2 = \bm{ijk} = -1$, and $\bm{ij}=\bm{k}$,  $\bm{jk}=\bm{i}$,  $\bm{ki}=\bm{j}$, $\bm{ji}=\bm{-k}$, $\bm{kj}=\bm{-i}$, $\bm{ik}=\bm{-j}$. The QNN uses the pure quaternion-valued feature $\bm{x_q} $ in the processing module, with the input feature $a$ and two fooling counterparts $b$ and $c$, which can be written as $\bm{x_q} = 0+a\bm{i}+b\bm{j}+c\bm{k}$. The encryption of $\bm{x_q}$ is to rotate it along a random axis $\bm{o}=0+o_1\bm{i}+o_2\bm{j}+o_3\bm{k}$ with a random rotation angle $\theta$, where $o_1, o_2, o_3 \in \mathbb{R}$ and $||\bm{o}||=1$. Thus, the encrypted feature can be described as $\bm{f_q} = \bm{R_q}\circ\bm{x_q}\circ\bm{\overline{R_q}}$, where $\bm{R_q}=e^{\bm{o}\frac{\theta}{2}}=\text{cos}\frac{\theta}{2}+\text{sin}\frac{\theta}{2}(o_1\bm{i}+o_2\bm{j}+o_3\bm{k})$, and $\bm{\overline{R_q}}=\textit{e}^{-\bm{o}\frac{\theta}{2}}$ is the conjugation of $\bm{R_q}$. Please see supplementary materials for more details.

As discussed above, we can directly transfer parameters from a well-trained complex-valued NN to the RENN with $d=2$. More crucially, this RENN ($d=2$) can also deal with $d'$-ary features under specific conditions when $d'>d$. It is because the $d'$-ary feature can be written analogous to the $d'$-ary hypercomplex number as $\bm{p}= a+b_1\bm{i}_1+\cdots+b_{(d'-1)}\bm{i}_{d'-1} \in \mathbb{H}_{d'}^{n}$, where $\bm{i}_m$ denotes the $m$-th imaginary part of the $(d'-1)$-ary number. In fact, the encryption in the scenario of using $2$-ary features can be considered as a special case as encryption in the scenario of $d'$-ary features when we use $a$ as the input feature and $b_1$ as the fooling counterpart. The other $(d'-2)$-ary features in $\bm{p}$ are set to zero, $b_2=b_3=\cdots=b_{(d'-1)}=\bm{0} \in \mathbb{R}^{n}$. Meanwhile, the rotation matrix $\bm{R} \in \mathbb{R}^{d' \times d'}$ is constrained as follows: $\bm{R}_{11}= \cos\theta, \bm{R}_{12}= \sin\theta, \bm{R}_{21}= -\sin\theta, \bm{R}_{22} = \cos\theta$; $\bm{R}_{ij}=0$ if $i \geqslant 2$ or $ j \geqslant 2$. In this way, the encryption based on $d'$-ary features is exactly equivariant to the encryptyion of $2$-ary features. Please supplementary materials for the proof.

In addition, RENNs usually have more potential phases to hide the input information than complex-valued NNs, because complex-valued NNs only have a single fooling counterpart. Experimental results have verified that RENNs have superior performance to complex-valued NNs.

\subsection{Attackers to RENNs}
\label{attacks}
In order to test the privacy protection performance of RENNs, we use\footnote[1]{Attackers can be directly extended to RENNs without revisions.} six different attackers designed in~\cite{deepPrivacy} to attack RENNs, which belong to the following two types.

$\bullet\quad$\textbf{Feature inversion attackers:}  Feature inversion attackers usually train another neural network to reconstruct the input using intermediate-layer features. There exist two feature inversion attackers. \textbf{Inversion attacker 1} estimates the rotation matrix $\bm{R}’$ encoding the input information, and uses the feature decrypted with $\bm{R}’$ to reconstruct the input $\hat{I} = \text{dec}_1(\Gamma({\bm{R}'}^{-1}\bm{f}))$. \textbf{Inversion attacker 2} directly uses the encrypted features to reconstruct the input, \emph{i.e.} $\hat{I} = \text{dec}_2(\bm{f})$. $\text{dec}_{1}(\cdot)$ and $\text{dec}_{2}(\cdot)$ denote the neural network trained for feature inversion attackers 1 and 2, respectively.

$\bullet\quad$\textbf{Property inference attackers:} The ``property'' refers to the input attribute~\cite{Property2018Karan}, and the inference attacker uses the intermediate-layer feature to infer sensitive attributes of inputs. In this paper, there are four types of inference attackers. \textbf{Inference attacker 1} trains the classifier to predict sensitive attributes using real images $I$ and ground-truth attributes $attr$, \emph{i.e.} $attr=net_1(I)$. During the attacking process, the attacker uses images reconstructed by the inversion attacker 1 to hack the sensitive attribute. \textbf{Inference attacker 2} decrypts the encrypted feature $\bm f$ with $\bm{R}’$ obtained by the \textbf{inversion attacker 1} to get the decrypted feature $a’$. The classifier is trained by the attacker using the decrypted feature $a'$ and ground-truth attributes $attr$, \emph{i.e.} $attr=net_2(a’)$. During the attacking process, the attacker uses the trained classifier and decrypted feature $a’$ to hack sensitive attributes. \textbf{Inference attacker 3} uses images $\hat{I}=\text{dec}_1(a’)$ reconstructed by the \textbf{inversion attacker 1} to train the classifier to predict sensitive attributes $attr$, \emph{i.e.} $attr=net_3(\hat{I})$. \textbf{Inference attacker 4} applies the $k$-nearest neighbors ($k$-NNs) instead of a neural network to infer sensitive attributes. Images, whose decrypted features $a’$ are close in the feature space, are considered to share similar attributes.

\section{Experiments}
In this section, we conducted a series of experiments to test  RENNs. Theoretically, the RENN is capable of being applied to different tasks. Whereas, considering the limitation of paper length, we converted a number of classical neural networks into RENNs for object classification and face attribute estimation, considering both the feature inversion attackers and the property inference attackers.

We trained RENNs using the CIFAR-10, CIFAR-100 datasets~\cite{cifar} with small images for object classification. Besides, the CUB200-2011 dataset~\cite{CUB200} and the CelebA dataset~\cite{CelebA} were used for object classification and face attributes estimation with large images, respectively. LeNet~\cite{lenet}, residual networks~\cite{resnet}, VGG-16~\cite{vgg}, and AlexNet~\cite{alexnet} were chosen to be revised into RENNs.

\textbf{Network architectures:}
Fig.~\ref{fig:structure} (right) compares architectures between the original neural network and its corresponding RENN.
The encoder/processing/decoder modules of each original neural network are introduced as follows. For the LeNet, the encoder consisted of all layers before the second convolutional layer, and there was only a softmax-layer in the decoder. For the residual network, the encoder was made up of all layers before the first $16\times16$ feature map, and the decoder contained all layers after the first $8\times8$ feature map. For the AlexNet, the encoder was composed of the first convolutional layers, and the decoder included fully-connected layers and the softmax layer. For the VGG-16, all layers before the last $56\times 56$ feature map comprised the encoder, and the decoder was composed of fully-connected layers and the softmax layer. Note that for each encoder, we added the GAN at the end of the encoder.

\textbf{Baselines:}
We proposed four baselines for comparison.
As shown in the second row of Fig.~\ref{fig:structure} (right), we used the original network without any revision as the first baseline, and denoted it as \emph{Original DNN}. The \emph{Original DNN} was divided in the same way as RENN into encoder, processing module, and decoder. The second baseline is shown in the third row. Considering noise addition was also helpful for privacy protection, we added noises to the output $a$ of the encoder, \emph{i.e.} $a\gamma\epsilon$, where $\epsilon$ denoted a random noise vector and $\gamma$ was a scalar. Hence, the second baseline was described as \emph{Noisy DNN}, and trained with $\gamma=0.2,0.5,1.0$. The third baseline is presented in the last row of Fig.~\ref{fig:structure} (right), which is termed \emph{``w/ additional layers"}. Since the insertion of the GAN increased the layer number of the RENN, for a fair comparison, the GAN architecture was regraded as the baseline network as well. Whereas, \emph{w/ additional layers} was learned without the GAN loss. \cite{deepPrivacy} was considered as the forth baseline and was simplified as \emph{Complex NN}, which had the same division of modules as the RENN.

\begin{table*}[!t]\vspace{-10pt}
\caption{Classification error rates and reconstruction errors indicating capacity of privacy protection.}
\label{tab:resnet_accuracy}\vspace{-8pt}
\begin{center}\resizebox{\linewidth}{!}{\
\begin{tabular}{c|cc|ccccc|c|cccccccc}
\toprule
\multirow{6}*{\rotatebox{90}{\tabincell{c}{Classification\\ Error Rate(\%)}}} \!\!\!\!&\!\!\!\! Model &\!\!\!\!\!\! Dataset &\!\!\!\! \tabincell{c}{Original\\DNN} \!\!\!\!&\!\!\!\! \tabincell{c}{w/ additional\\ layers} \!\!\!\!&\!\!\!\! \tabincell{c}{Complex\\NN} \!\!\!\!&\!\!\!\! \tabincell{c}{RENN\\$d=3$}\!\!\!\!&\!\!\!\! \tabincell{c}{RENN\\$d=5$}
\!\!\!\!&\!\!\!\! \multirow{6}*{\rotatebox{90}{\tabincell{c}{Reconstruction\\ Errors}}} \!\!\!\!&\!\!\!\!\tabincell{c}{Original\\DNN} \!\!\!\!&\!\!\!\! \tabincell{c}{w/ additional\\ layers} \!\!\!\!&\!\!\!\! \tabincell{c}{Complex\\ dec($a'$)} \!\!\!\!&\!\!\!\! \tabincell{c}{Complex\\ dec($x$)} \!\!\!\!&\!\!\!\! \tabincell{c}{RENN($d=3$)\\ dec($a'$)} \!\!\!\!&\!\!\!\! \tabincell{c}{RENN($d=3$)\\ dec($x$)} \!\!\!\!&\!\!\!\! \tabincell{c}{RENN($d=5$)\\ dec($a'$)} \!\!\!\!&\!\!\!\! \tabincell{c}{RENN($d=5$)\\ dec($x$)}  \\
\cmidrule{2-8}\cmidrule{10-17}
 &
ResNet-20 &\!\!\!\!\!\! CIFAR-10 & 11.56 & 9.68 & 10.91 & \textbf{9.21} &- & & 0.0906 & 0.1225 & 0.2664 & 0.2420 & \textbf{0.3014} & 0.2702 & - & - \\
&ResNet-32 &\!\!\!\!\!\! CIFAR-10 & 11.13 & 9.67 & 10.48 & \textbf{9.82} & -& & 0.0930 & 0.1171 & 0.2569 &
0.2412 & \textbf{0.2813} & 0.2412 & - & -\\
&ResNet-44 &\!\!\!\!\!\! CIFAR-10 & 10.67 & 9.43 & 11.08 & \textbf{9.54} & -& & 0.0933 & 0.1109 & 0.2746 &
0.2419 & \textbf{0.3123} & 0.2421& - & - \\
&ResNet-56 &\!\!\!\!\!\! CIFAR-10 & 10.17 & 9.16 & 11.53 & 9.24 & \textbf{7.73} & & 0.0989 & 0.1304 & 0.2804 & 0.2377 & 0.3083 & 0.2403 & \textbf{0.3285} & 0.2542 \\
&ResNet-110 &\!\!\!\!\!\! CIFAR-10 & 10.19 & 9.14 & 11.97 & \textbf{9.31} &- & & 0.0896 & 0.1079 & \textbf{0.3081} &
0.2495 & 0.3028 & 0.2379 & - & -\\
\bottomrule
\end{tabular}
}
\resizebox{\linewidth}{!}{\
\begin{tabular}{c|cc|cc|ccc|cccccc}
\toprule
\multirow{7}*{\rotatebox{90}{\tabincell{c}{Reconstruction\\ Errors}}} \!\!\!\!&\!\!\!\! Model \!\!\!\!&\!\!\!\! Dataset \!\!\!\!&\!\!\!\! \tabincell{c}{Original\\DNN} \!\!\!\!&\!\!\!\! \tabincell{c}{w/additional\\ layers} \!\!\!\!&\!\!\!\! \tabincell{c}{Noisy DNN\\$\gamma=0.2$} \!\!\!\!&\!\!\!\! \tabincell{c}{Noisy DNN\\$\gamma=0.5$}
 \!\!\!\!&\!\!\!\! \tabincell{c}{Noisy DNN\\$\gamma=1.0$} \!\!\!\!&\!\!\!\! \tabincell{c}{Complex\\dec($a'$)} \!\!\!\!&\!\!\!\!
\tabincell{c}{Complex\\dec($x$)} \!\!\!\!&\!\!\!\! \tabincell{c}{RENN ($d=3$)\\dec($a'$)} \!\!\!\!&\!\!\!\!
\tabincell{c}{RENN ($d=3$)\\dec($x$)} \!\!\!\!&\!\!\!\! \tabincell{c}{RENN ($d=5$)\\dec($a'$)} \!\!\!\!&\!\!\!\!
\tabincell{c}{RENN ($d=5$)\\dec($x$)} \\
\cmidrule{2-14}
& LeNet&\!\!\!\! CIFAR-10 \!\!\!\!& 0.0769 & 0.1208 & 0.0948 & 0.1076 & 0.1274 & 0.2405 & 0.2353 & 0.2877 & 0.2303 & \textbf{0.3021} &0.2577\\
& LeNet&\!\!\!\! CIFAR-100 \!\!\!\!& 0.0708 & 0.1314 & 0.0950 & 0.1012 & 0.1286 & 0.2700 & 0.2483 & \textbf{0.2996} & 0.2528 & - & -\\
& ResNet-56 &\!\!\!\! CIFAR-100 \!\!\!\!& 0.0929 & 0.1029 & 0.1461 & 0.1691 & 0.2017 & 0.2593 & 0.2473 & \textbf{0.3057} & 0.2592 & - & -\\
& ResNet-110 &\!\!\!\! CIFAR-100 \!\!\!\!& 0.1050 & 0.1092 & 0.1483 & 0.1690 & 0.2116 & 0.2602 & 0.2419 & \textbf{0.3019} & 0.2543 & - & -\\
& VGG-16 &\!\!\!\! CUB200-2011 \!\!\!\!& 0.1285 & 0.1202 & 0.1764 & 0.0972 & 0.1990 & 0.2803 & 0.2100 & \textbf{0.3133} & 0.1945 & - & - \\
& AlexNet &\!\!\!\! CelebA \!\!\!\!& 0.0687 & 0.1068 & - & - & - & 0.3272 & 0.2597 & 0.3239 & 0.2657 & \textbf{0.3432} & 0.2766 \\

\bottomrule
\end{tabular}
}\end{center}\vskip -0.1in\vspace{-10pt}
\end{table*}

\begin{figure*}[t]
\begin{center}
\centerline{\includegraphics[width=0.75\linewidth]{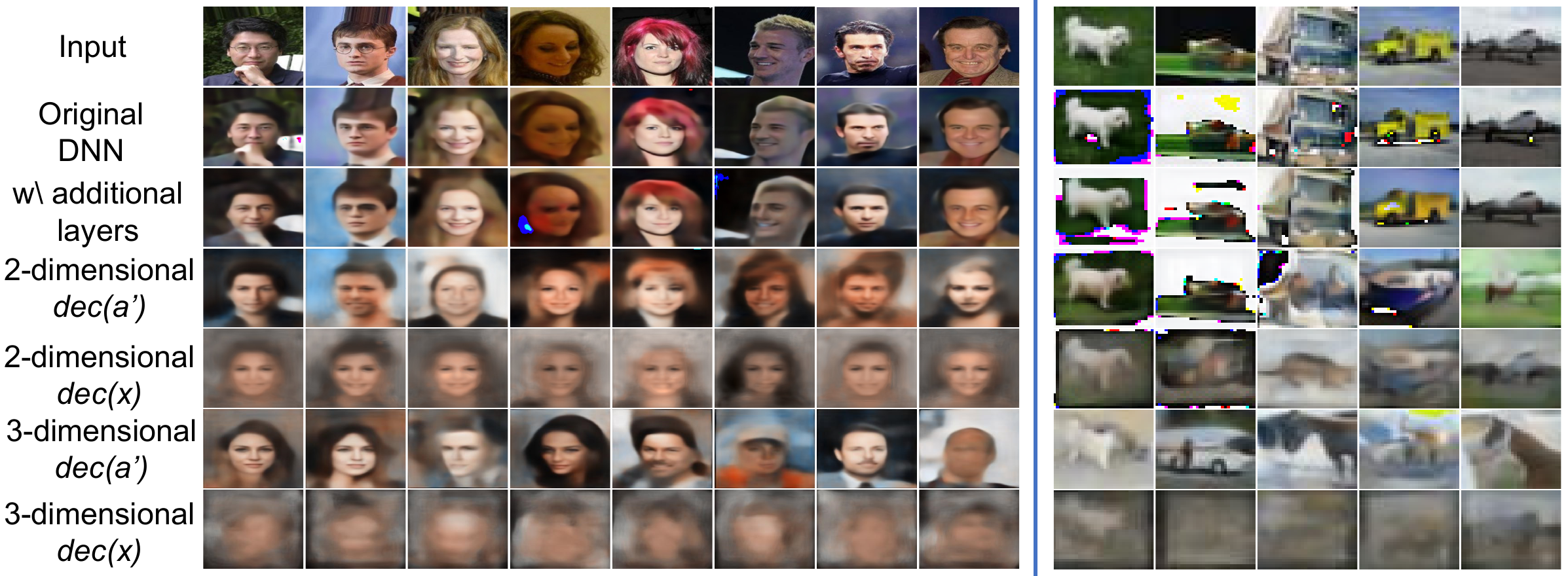}}
\vspace{-5pt}
\caption{Images reconstructed from features. Left: images from the CelebA dataset; right: images from the CIFAR-10 dataset. Please see supplementary materials for more results.}
\label{fig:reconstruct}\end{center}\vskip -0.2in\vspace{-5pt}\end{figure*}

\begin{table}[t!]
\vspace{-5pt}
\begin{minipage}[h]{0.58\linewidth}
\begin{center}
\caption{Classification error rates on various models and datasets.}\label{tab:other_accuracy}
\resizebox{\linewidth}{!}{\
\begin{small}
 \begin{tabular}{c|cc|ccc|ccc}
  \toprule
  Model, Dataset \!\!\!\!&\!\!\tabincell{c}{ Original\\DNN} \!\!\!\!\!&\!\!\!\! \tabincell{c}{w/ additional\\ layers}  \!\!\!\!&\!\!\!\!  \tabincell{c}{Noisy DNN\\$\gamma=0.2$}  \!\!\!\!\!&\!\!\!\!  \tabincell{c}{Noisy DNN\\$\gamma=0.5$}
    \!\!\!\!\!&\!\!\!\!  \tabincell{c}{Noisy DNN\\$\gamma=1.0$}  \!\!\!\!&\!\!\!\!  \tabincell{c}{Complex\\NN}  \!\!\!\!\!&\!\!\!\!  \tabincell{c}{RENN\\$d=3$}  \!\!\!\!\!&\!\!\!\!  \tabincell{c}{RENN\\$d=5$} \\
  \midrule
  LeNet, CIFAR-10 \!\!\!\!\!&\!\!\!\!\! 19.78 \!\!\!\!\!&\!\!\!\!\! 21.52 \!\!\!\!\!&\!\!\!\!\! 24.15 \!\!\!\!\!&\!\!\!\!\! 27.53 \!\!\!\!\!&\!\!\!\!\! 34.43 \!\!\!\!\!&\!\!\!\!\! 17.95  \!\!\!\!\!&\!\!\!\!\!  11.45 \!\!\!\!\!&\!\!\!\!\! \textbf{11.39}\\
  LeNet, CIFAR-100 \!\!\!\!\!&\!\!\!\!\! 51.45 \!\!\!\!\!&\!\!\!\!\! 49.85 \!\!\!\!\!&\!\!\!\!\! 56.65 \!\!\!\!\!&\!\!\!\!\! 67.66 \!\!\!\!\!&\!\!\!\!\! 78.82 \!\!\!\!\!&\!\!\!\!\! 49.76  \!\!\!\!\!&\!\!\!\!\!  \textbf{37.78} \!\!\!\!\!&\!\!\!\!\! - \\
  ResNet-56, CIFAR-100 \!\!\!\!\!&\!\!\!\!\! 53.26 \!\!\!\!\!&\!\!\!\!\! 44.38 \!\!\!\!\!&\!\!\!\!\! 57.24 \!\!\!\!\!&\!\!\!\!\! 61.31 \!\!\!\!\!&\!\!\!\!\! 74.17 \!\!\!\!\!&\!\!\!\!\! \textbf{44.37}  \!\!\!\!\!&\!\!\!\!\!  44.86 \!\!\!\!\!&\!\!\!\!\! - \\
  ResNet-110, CIFAR-100 \!\!\!\!\!&\!\!\!\!\! 50.64 \!\!\!\!\!&\!\!\!\!\! 44.93 \!\!\!\!\!&\!\!\!\!\! 55.19 \!\!\!\!\!&\!\!\!\!\! 61.12 \!\!\!\!\!&\!\!\!\!\! 71.31 \!\!\!\!\!&\!\!\!\!\! 50.94  \!\!\!\!\!&\!\!\!\!\!  \textbf{42.05} \!\!\!\!\!&\!\!\!\!\! - \\
  VGG-16, CUB200-2011   \!\!\!\!\!&\!\!\!\!\! 56.78 \!\!\!\!\!&\!\!\!\!\! 63.47 \!\!\!\!\!&\!\!\!\!\! 69.20 \!\!\!\!\!&\!\!\!\!\! 99.48 \!\!\!\!\!&\!\!\!\!\! 99.48 \!\!\!\!\!&\!\!\!\!\! 78.50  \!\!\!\!\!&\!\!\!\!\!  \textbf{70.86} \!\!\!\!\!&\!\!\!\!\!- \\
  AlexNet, CelebA \!\!\!\!\!&\!\!\!\!\! 14.17 \!\!\!\!\!&\!\!\!\!\! 9.49 \!\!\!\!\!&\!\!\!\!\! - \!\!\!\!\!&\!\!\!\!\! - \!\!\!\!\!&\!\!\!\!\! - \!\!\!\!\!&\!\!\!\!\! 15.94  \!\!\!\!\!&\!\!\!\!\!  \textbf{8.80} \!\!\!\!\!&\!\!\!\!\! 9.23\\
  \bottomrule
  \end{tabular}
\end{small}
}
\end{center}
\end{minipage}\hfill
\begin{minipage}[h]{0.41\linewidth}
  \begin{center}
 \caption{Rank of the estimated sample, and time cost of inference.}
\label{tab:k-anonymity}
  \resizebox{\columnwidth}{!}{\
  \begin{small}
\begin{tabular}{c|ccccc}
  \toprule
  \!\!\!\!\!\!&\!\!\!\! \tabincell{c}{w/ additional\\ layers} \!\!\!\!&\!\!\!\!\tabincell{c}{Complex\\NN} \!\!\!\!&\!\!\!\!\tabincell{c}{RENN\\$d=3$}\!\!\!\!&\!\!\!\! \tabincell{c}{RENN\\$d=5$}\!\!\!\!&\!\!\!\! \tabincell{c}{Homomorphic\\encryption} \\
  \midrule
  Rank \!\!\!\!& - \!\!\!\!&\!\!\!\!\!\!\!\! 9.27 \!\!\!\!&\!\!\!\!\!\! 241.49 \!\!\!\!\!\!&\!\! 46265.12 & -\\
  \tabincell{c}{Time cost\\ (s/image)} \!\!\!\!& 0.0004 \!\!\!\!&\!\!\!\!\!\!\!\! 0.0007 \!\!\!\!&\!\!\!\! 0.0011 \!\!\!\!& \!\!\!\! 0.0044 \!\!\!\! & \!\!\!\! 3.56 \\
  \bottomrule
  \end{tabular}
  \end{small}
  }
  \end{center}
  \end{minipage}\vspace{-5pt}
\end{table}

\begin{table*}[!t]
\vspace{-5pt}
\caption{Failure rate of identifying the reconstructed image by human annotators.}
\label{tab:error_human}\vspace{-12pt}
\begin{center}
\resizebox{\linewidth}{!}{\
\begin{small}
\begin{tabular}{lc|cc|ccc|cccccc}
\toprule

Model \!\!\!\!&\!\!\!\! Dataset &\!\!\!\! \tabincell{c}{Original\\DNN} \!\!\!\!&\!\!\!\! \tabincell{c}{w/ additional\\ layers} \!\!\!\!&\!\!\!\! \tabincell{c}{Noisy DNN\\$\gamma=0.2$} \!\!\!\!&\!\!\!\! \tabincell{c}{Noisy DNN\\$\gamma=0.5$}
 \!\!\!\!&\!\!\!\! \tabincell{c}{Noisy DNN\\$\gamma=1.0$} \!\!\!\!&\!\!\!\! \tabincell{c}{Complex\\dec($a'$)} \!\!\!\!&\!\!\!\!
\tabincell{c}{Complex\\dec($x$)} \!\!\!\!&\!\!\!\! \tabincell{c}{RENN ($d=3$)\\dec($a'$)} \!\!\!\!&\!\!\!\!
\tabincell{c}{RENN ($d=3$)\\dec($x$)} \!\!\!\!&\!\!\!\! \tabincell{c}{RENN ($d=5$)\\dec($a'$)} \!\!\!\!&\!\!\!\!
\tabincell{c}{RENN ($d=5$)\\dec($x$)} \\
\midrule
LeNet & CIFAR-10 & 0.16 & 0.12 & 0.20 & 0.20 & 0.24 & 0.82 & 0.92 & 0.90 & 0.96 & 0.92 & \textbf{1.00}\\
LeNet & CIFAR-100 & 0.16 & 0.12 & 0.20 & 0.64 & 0.72 & 0.80 & 0.92 & 0.94 & \textbf{1.00} &- &-\\
ResNet-56 & CIFAR-100 & 0.06 & 0.06 & 0.08 & 0.10 & 0.36 & 0.72 & 0.88 & 0.90 & \textbf{1.00} &- &- \\
ResNEt-110 & CIFAR-100 & 0.04 & 0.12 & 0.10 & 0.16 & 0.36 & 0.80 & 0.86 & 0.94 & \textbf{0.98} &- &-\\
VGG-16 \!\!\!\!&\!\!\!\! CUB200-2011 & 0.06 & 0.06 & 0.08 & 0.02 & 0.14 & 0.86 & 0.84 & \textbf{0.86} & 0.74 &- &-\\
AlexNet & CelebA & 0.04 & 0.24 & - & - & - & 0.96 & 1.00 & 0.84 & \textbf{1.00}& 0.94 & \textbf{1.00}\\
\bottomrule
\end{tabular}
\end{small}
}
\end{center}\vspace{-18pt}\end{table*}

\textbf{Attackers:} We applied two kinds of attackers in Sec.~\ref{attacks}.

$\bullet\,$ \emph{Inversion attackers:}
The inversion attacker was implemented based on U-Net~\cite{UNet}. The intermediate-layer feature was upsampled to the size of the input, which was fed into the inversion model. There were four down-sample blocks and four up-sample blocks. Each block had six convolutional layers for better performance, and the output of the inversion model had the same size as the input. Intermediate-layer features from RENNs was used as the input of the attacker. In this way, the first inversion attacker and the second inversion attacker were constructed based on the U-Net, according to Sec.~\ref{attacks}. The first inversion attacker was trained with $d$-ary features, and tested with the output of $D$. The second inversion attack trained and tested the attacker with the encrypted $d$-ary features. To mimic the procedure of hacking the privacy, we randomly sampled the rotation matrix for 1000 times. The sample with the highest output of the discriminator was considered as the optimal feature to reconstruct the original image.

$\bullet\,$ \emph{Inference attackers:}
For the inference attack, we used the CelebA and CIFAR-100 datasets for testing. 10 attributes were selected as private attributes from the CelebA dataset. An AlexNet was revised to a RENN, which was trained to estimate other 30 attributes. Whereas the attacker based on ResNet-50 used the intermediate-layer feature to estimate private attributes. For the CIFAR-100 dataset, we transformed a ResNet-56 into a RENN to classify major 20 superclasses of CIFAR-100. The attacker based on ResNet-56 used the intermediate-layer feature to infer 100 minor classes, which were considered as sensitive information in this experiment.

\begin{table}[t]
\vspace{-5pt}
\begin{minipage}[h]{0.50\linewidth}
\begin{center}
\caption{Average error of the estimated angle.}
\vspace{10pt}
\label{tab:error_angle}
\resizebox{\linewidth}{!}{\
\begin{small}
\begin{tabular}{lc|ccc}
\toprule
& & \multicolumn{2}{c}{Average Error} \\
\midrule
Model & Dataset & \tabincell{c}{Error of \\ Complex NN} & \tabincell{c}{Error of\\ RENN ($d=3$)} & \tabincell{c}{Error of\\ RENN ($d=5$)}\\
\midrule
ResNet-20 & CIFAR-10 & 0.7890$\pm$0.3722 & 1.4803$\pm$0.7004 & -\\
ResNet-32 & CIFAR-10 & 0.7820$\pm$0.3630 & 1.3322$\pm$0.6785 & -\\
ResNet-44 & CIFAR-10 & 0.8411$\pm$0.6200 & 1.4610$\pm$0.6905 & -\\
ResNet-56 & CIFAR-10 & 0.8088$\pm$0.5848 & 1.4733$\pm$0.6932 & 1.5802$\pm$0.4927\\
ResNet-110 & CIFAR-10 & 0.8048$\pm$0.4535 & 1.4461$\pm$0.6955 & -\\
LeNet & CIFAR-10 & 0.7884$\pm$0.4147 & 1.3511$\pm$0.6765 & 1.5355$\pm$0.5043\\
LeNet & CIFAR-100 & 0.8046$\pm$0.5279 & 1.3842$\pm$0.6694  & -\\
ResNet-56 & CIFAR-100 & 0.7898$\pm$0.5544 & 1.3837$\pm$0.7010 & -\\
ResNet-110 & CIFAR-100 & 0.7878$\pm$0.3775 & 1.3515$\pm$0.6558 & -\\
VGG-16 \!\!\!\!\!\!\!\!&\!\!\!\!\!\!\!\! CUB200 \!\!\!\!\!\!\!\!\!\!&\!\!\!\! 1.5572$\pm$0.8778 & 1.3589$\pm$0.7120 & -\\
AlexNet & CelebA & 0.8500$\pm$0.5811 & 1.3833$\pm$0.6767 & 1.5571$\pm$0.4959 \\
\bottomrule
\end{tabular}
\end{small}
}
\end{center}
\end{minipage}\hfill
\begin{minipage}[h]{0.48\linewidth}
  \begin{center}
  \caption{Experimental results of inference attackers. net($I$), net($a'$), net($\hat{I}$), $k$-NN represented the first, the second, the third, and the forth inference attacker, respectively.}
  \label{tab:inference_error}
  \resizebox{\columnwidth}{!}{\
  \begin{small}
  \begin{tabular}{c|c|c|c|c|c|ccc}
  \toprule
  & &\!\!\!\! \tabincell{c}{Classification\\Error Rate} \!\!\!\!&\!\!\!\!
  net($I$) \!\!\!\!&\!\!\!\! net($a'$) \!\!\!\!&\!\!\!\! net($\hat{I}$) \!\!\!\!&  \multicolumn{3}{c}{$k$-NN} \\
  \midrule
   & Structure & & & & &\!\!\!\! $k=1$ \!\!\!\!&\!\!\!\! $k=3$ \!\!\!\!&\!\!\!\! $k=5$ \!\!\!\!\\
  \midrule
  \multirow{3}*{\rotatebox{90}{\tabincell{c}{CIFAR-\\100}}} \!\!\!\!\!\!&\!\!\!\! w/ additional layers \!\!\!\!& 18.73 & 68.72 & 38.50 & 40.28 & 73.38 & 68.37 & 71.16\\
   & Complex NN & 26.77 & 94.53 & 87.17 & 89.56 & 94.44 & 93.63 & 92.50 \\
   & RENN ($d=3$) & 21.60 & 96.73 & 94.22 & 95.25 & 98.19 & 97.60 & 97.42 \\
   & RENN ($d=5$) & 22.33 & \textbf{97.92} & \textbf{98.14} & \textbf{98.22} & \textbf{98.67} & \textbf{98.56} & \textbf{98.48} \\
  \midrule
  \multirow{3}*{\rotatebox{90}{CelebA}} \!\!\!\!\!\!&\!\!\!\! w/ additional layers \!\!\!\!& 8.04 & 19.14 & 13.17 & 14.01 & 20.14 & 17.26 & 16.20 \\
   & Complex NN & 14.75 & \textbf{25.72} & 22.21 & 22.61 & 31.69 & 27.90 & 26.41 \\
   & RENN ($d=3$) & 8.20 & 25.03 & 22.19 & 23.26 & \textbf{32.77} & 28.81 & 27.42\\
   & RENN ($d=5$) & 7.95 & 25.41 & \textbf{25.69} & \textbf{25.53} & 32.63 & \textbf{28.95} & \textbf{27.55}\\
  \bottomrule
  \end{tabular}
  \end{small}
  }
  \end{center}
  \end{minipage}
\end{table}

\begin{figure}[t]
\vspace{-8pt}
\begin{center}
\includegraphics[width=0.8\linewidth]{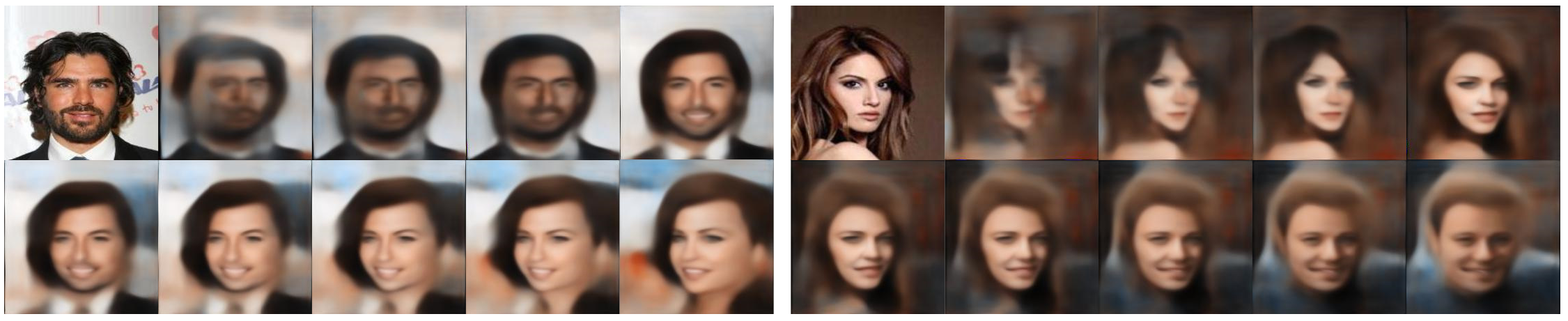}\vspace{-5pt}
\caption{CelebA images reconstructed using different phases. The first image in each group is the input image. We show the most meaningful images. More results are shown in supplementary materials.}
\label{fig:k-anonymity}\end{center}\vskip -0.2in \vspace{-8pt} \end{figure}

\textbf{Evaluation metrics of privacy protection:}
We used four metrics to evaluate the performance of privacy protection in terms of inversion attackers: (1) the pixel-level reconstruction error $\mathbb{E}[\|\hat{I} - I\|]$, where the pixel value was scaled to $[0, 1]$. (2) The average difference angle $\Delta \theta$ between the decrypted input information $a$ and the decrypted feature $a'$ by the attacker,~\emph{i.e.} $\Delta \theta = \arccos \langle\bm{R}\cdot\rho, \bm{R'}\cdot \rho \rangle$, where $\rho=[1,0,0,\cdots] ^T\in \mathbb{R}^d$ (3) The rank of the estimated sample. Let us recover two samples using two similar phases. According to our experience, when the angle between phases was less than $\pi/36$, two samples usually presented the same object entity. Thus, we estimated the number of object entities that were more similar to the input than the recovered sample. For the $5$-ary RENN, $3$-ary RENN and Complex NN, ranks of the estimated sample are computed as $\frac{4}{2-3\cos\Delta\theta+ \cos^{3}\Delta\theta}$, $\frac{2}{1-\cos\Delta\theta}$ and $\frac{2\pi}{\Delta\theta}$, respectively. Please see supplementary materials for derivations of above formulas. (4) The reconstruction failure rate of human identification, \emph{i.e.} we used human annotators to judge whether they can identify the input based on the reconstructed sample $\hat{I}$. For the privacy performance of inference attacks, the accuracy of attackers can be regarded as the evaluation metric. Moreover, the processing speed of different models can be computed to measure the efficiency of RENNs.

\textbf{Experimental Results and Analysis:}
\label{results}
Table~\ref{tab:resnet_accuracy} and Table~\ref{tab:other_accuracy} show the performance of RENNs and baselines. Complex $dec(a')$ and Complex $dec(x)$ represented the inversion attacker $1$ and the inversion attacker $2$ on Complex NN, respectively. ``RENN ($d=3$) $dec(a')$,'' ``RENN ($d=3$) $dec(x)$,'' ``RENN ($d=5$) $dec(a')$,'' ``RENN ($d=5$) $dec(x)$'' represented the inversion attacker $1$ based on $dec(a')$ and the inversion attacker $2$ based on $dec(x)$ on the $3$-ary RENN and the $5$-ary RENN, respectively. Classification error rate results showed that RENNs with $d=3$ and RENNs with $d=5$ achieved better performances than Complex NNs. Compared with Original DNN and DNNs with additional layers, the accuracy of RENNs ($d=3$) and RENNs ($d=5$) was not significantly affected. As for the reconstruction error, a higher value indicated a better privacy protection performance. Reconstruction errors of $5$-ary RENNs were higher than other networks, \emph{i.e.} $5$-ary RENNs exhibited better performance of privacy protection than other networks.

Fig.~\ref{fig:reconstruct} visualizes several examples from inversion attackers. We only provided results from partial experiments constrained by the space. Please see supplementary materials for more results. Table~\ref{tab:error_angle} shows averages and standard errors of $\Delta \theta$. A smaller value of the average $\Delta \theta$ indicated that attackers were easier to estimate the target phase. Table~\ref{tab:error_human} shows the subjective failure rate according to the judgement of humans. Attackers could not fetch the input information from encrypted features. It was more difficult to estimate the target phase of $3$-ary and $5$-ary features than complex-valued features.

Fig.~\ref{fig:k-anonymity} shows images reconstructed using different phases, and Table~\ref{tab:k-anonymity} shows the rank of the estimated sample. A higher rank value indicated better performance of privacy protection. The ranks of the $3$-ary RENN and the $5$-ary RENN were higher than the Complex NN. \emph{I.e.} it was more difficult for attackers on RENNs to find the target phase than attackers on Complex NNs.

Table~\ref{tab:k-anonymity} also shows the time cost of the inference process of DNNs. The time cost of homomorphic encryption was from the framework Gazelle~\cite{Gazelle}, which trained a small network with 3 fully-connected layers from~\cite{SecureML} using the CIFAR-10 dataset. For other networks, we used networks revised from the ResNet-56, which were deeper than the network used by Gazelle. However, the inference time cost of the $3$-ary RENN was much less than Gazelle, and was comparable with traditional DNNs.

Table~\ref{tab:inference_error} shows the result of inference attackers. A higher value of inference error indicated a better privacy protection performance. Attackers on $5$-ary RENNs had a higher inference classification error than attackers on DNNs with additional layers, and attackers on Complex NN. Thus, $5$-ary RENNs protected private attributes from attackers more effectively.

\section{Conclusion}
In this paper, we propose a method to protect the privacy of inputs. Our method transforms traditional DNNs into RENNs, which use $d$-ary features as intermediate-layer features. The input information is hidden in a random phase of $d$-ary features. Experiments showed the effectiveness of the privacy protection of RENNs, which have much lower computational cost than the homomorphic encryption.
\section*{Broader Impact}
This study has broad impacts on privacy protection in DNNs. Our research provides researchers with a set of generic rules to revise various traditional DNNs into rotation-equivariant neural networks for privacy protection. Compared to homomorphic encryption, our RENN requires significantly less computational cost. Crucially, the complex-valued NN can be considered as a specific case of a RENN. The superior performance and the generality of the theory ensure that the RENN has broad impacts on both theory and practice.

\bibliography{privacy}
\bibliographystyle{plainnat}

\newpage
\appendix
\section{Rotation Equivariance}
\subsection{Applying the convolution operation to a $d$-ary feature is equivalent to applying the same convolution operation to each component in the $d$-ary feature}
In the Sec. 2.1 of the paper, we briefly introduce the revision of the convolution operation. In this section, we aim to show that applying the convolution operation to a $d$-ary feature is equivalent to applying the same convolution operation to each component in the $d$-ary feature.

Given a $d$-ary feature ${\bm{f}}=[a,b_1,b_2,\cdots,b_{d-1}]\in\mathbb{H}_d^n$ and a real-valued vector $w\in\mathbb{R}^n$, as discussed in the paper, the $d$-ary feature $\bm{f}$ can also be represented as a $n\times d$ matrix.
\begin{equation}
\label{eqn:A1}
{\bm{f}}=\begin{bmatrix}
  a_1&b_{1,1}&b_{2,1}&\cdots&b_{d-1,1}\\
  a_2&b_{1,2}&b_{2,2}&\cdots&b_{d-1,2}\\
  \vdots& \vdots& \vdots& \ddots& \vdots\\
  a_n& b_{1,n}& b_{2,n}&\cdots&b_{d-1,n}
\end{bmatrix}.
\end{equation}
Then, we have
\begin{equation}
\label{eqn:A2}
\begin{aligned}
w^\text{T}{\bm{f}}&=
[w_1,w_2,\cdots,w_n]
\times
\begin{bmatrix}
  a_1&b_{1,1}&b_{2,1}&\cdots&b_{d-1,1}\\
  a_2&b_{1,2}&b_{2,2}&\cdots&b_{d-1,2}\\
  \vdots& \vdots& \vdots& \ddots& \vdots\\
  a_n& b_{1,n}& b_{2,n}&\cdots&b_{d-1,n}
\end{bmatrix}
\\
&=\left[\sum_{i=1}^na_iw_i,
\sum_{i=1}^nb_{1,i}w_i,
\cdots,
\sum_{i=1}^nb_{d-1,i}w_i
\right]\\
&=\left[w^\text{T}a,w^\text{T}b_1,w^\text{T}b_2,\cdots,w^\text{T}b_{d-1}\right]
\end{aligned}
\end{equation}
Thus, applying a convolution operation to the $d$-ary feature is equivalent to applying the convolution operation to each component of the $d$-ary feature.

\subsection{Signal processing of a $d$-ary feature is equivalent to applying the same signal processing to each component in the $d$-ary feature}
Assuming that $\bm{f = x \cdot R}$, the output of the processing module ${\Phi}(\cdot)$ using ReLU as non-linear layers can be written as
\begin{equation}
\label{eqn:A3}
\begin{aligned}
{\Phi}({\bm{f}})&=\Phi_L(\Phi_{L-1}(\cdots\Phi_1(\bm{x \cdot R})))\\
&=\sigma(W_L^\text{T}\sigma(W_{L-1}^\text{T}(\cdots\sigma(W_1^\text{T}{\bm{x \cdot R}}))))
\end{aligned}
\end{equation}
where $\sigma(\cdot)$ represent the function of ReLU. The ReLU function can be considered as the element-wise multiplication. For the first ReLU layer in the processing module, we can represent the ReLU function as $\sigma(W_1^\text{T}\bm{f}) = \Sigma_1W_1^\text{T}\bm{f}$, where $\Sigma_1= \text{diag}(c_1; c_2;c_3;\cdots;c_n)$. $c_i \in \{0,1\}$ denotes the binary gating state for the $i$-th element in $W_1^\text{T}\bm{f}$, and $n$ is the number of elements in $W_1^\text{T}\bm{f}$. In this way, a real valued matrix $A=\Sigma_LW_L^\text{T}\Sigma_{L-1}W_{L-1}^\text{T}\cdots\Sigma_1W_1^\text{T}$ denotes the effect that combines all transformations in ${\Phi}({\bm{f}})$.

Thus, we can rewrite the above equation as follows
\begin{equation}
\label{eqn:A4}
\begin{aligned}
{\Phi}({\bm{f}})&=\sigma(W_L^\text{T}\sigma(W_{L-1}^\text{T}(\cdots\sigma(W_1^\text{T}{\bm{f}}))))\\
&=\Sigma_LW_L^\text{T}\Sigma_{L-1}W_{L-1}^\text{T}\cdots\Sigma_1W_1^\text{T}{\bm{f}}\\
&=A{\bm{f}}
\end{aligned}
\end{equation}
According to Eqn.~\eqref{eqn:A2} and Eqn.~\eqref{eqn:A4}, we have
\begin{equation}
\label{eqn:A5}
\begin{aligned}
{\Phi}({\bm{f}})&=A{\bm{f}}\\
&=[A\bm{f}^1,A\bm{f}^2,\cdots,A\bm{f}^{d}]\\
&=[{\Phi}(\bm{f}^1),{\Phi}(\bm{f}^2),\cdots,{\Phi}(\bm{f}^{d})].
\end{aligned}
\end{equation}
where $\bm{f}^i$ denotes the $i$-th component of the $d$-ary feature $\bm{f}$. Therefore, the signal processing of  $\bm{f}$ by the processing module is equivalent to applying the same signal processing to each component of $\bm{f}$.

\subsection{Proof of $\Phi({{\bf f} \cdot {\bf R}})=\Phi({\bf f})\cdot{\bf R}$}
In this paper, we revise the operation of each layer in the processing module to ensure that the input information is always encoded in the same phase of all $d$-ary features. Recursively, layerwise operations in processing module are supposed to satisfy
$$\Phi(\bm{f \cdot R})=\Phi(\bm{f})\cdot\bm{R}$$
Where $\bm{\Phi(\cdot)}$ is a certain layerwise operation and $\bm{f}\in \mathbb{H}_{d}^{n}$ is a real-valued intermediate-layer feature\\
Let us consider the following six most common types of layers/operations that are revised to construct the processing module, \textit{i}.\textit{e}. convolutional layer, ReLU, batch-normalization, Avg/Max-pooling, drop-out and skip-connection in Sec. 2.1 of the paper.
\subsubsection{Convolutional layer (or fully-connected layer)}
For revised convolutional layer, we remove bias term. Thus, we get
$$\begin{aligned}
\text{Conv}(\bm{f})&= w \cdot \bm{f}\\
&=\left[
\begin{matrix}
w_{11} & \cdots & w_{1n}\\
\vdots & \ddots & \vdots\\
w_{D1} & \cdots & w_{Dn}\\
\end{matrix}
\right]
\cdot
\left[
\begin{matrix}
\bm{f}_1\\
\vdots\\
\bm{f}_n\\
\end{matrix}
\right]\\
&=\left[
\begin{matrix}
(w_{11}\bm{f}_1+\cdots+w_{1n}\bm{f}_n)\\
\vdots\\
(w_{D1}\bm{f}_1+\cdots+w_{Dn}\bm{f}_n)\\
\end{matrix}
\right]
\end{aligned}
$$
where $\bm{f}_v\in \mathbb{H}_{d}$ denotes the $v$-th ($1\leq v \leq n$) $d$-ary element in the feature $\bm{f} = [\bm{f}_1, \cdots, \bm{f}_n]^{T} \in \mathbb{H}_{d}^{n} $. Here, we represent $\bm{f}$ as a $n \times d$ matrix. Thus we get
$$
\begin{aligned}
\text{Conv}(\bm{f} \cdot \bm{R}) &=w \cdot(\bm{f}\cdot \bm{R}) \\
&=\left[
\begin{matrix}
  w_{11} & \cdots & w_{1n}\\
  \vdots & \ddots & \vdots\\
  w_{D1} & \cdots & w_{Dn}\\
\end{matrix}
\right]\cdot
(
\left[
\begin{matrix}
\bm{f}_1\\
\vdots\\
\bm{f}_n\\
\end{matrix}
\right]\cdot \bm{R} )\\
&=\left[
\begin{matrix}
(w_{11}(\bm{f}_1\bm{R})+\cdots+w_{1n}(\bm{f}_n \bm{R}))\\
\vdots\\
(w_{D1}(\bm{f}_1 \bm{R})+\cdots+w_{Dn}(\bm{f}_n \bm{R}))\\
\end{matrix}
\right]\\
&=\left[
\begin{matrix}
((w_{11}\bm{f}_1 \bm{R})+\cdots+(w_{1n}\bm{f}_n \bm{R}))\\
\vdots\\
((w_{D1}\bm{f}_1 \bm{R})+\cdots+(w_{Dn}\bm{f}_n \bm{R}))\\
\end{matrix}
\right]\\
&=(w  \bm{f} )\cdot \bm{R} = \text{Conv}(\bm{f}) \cdot \bm{R}
\end{aligned}
$$

\subsubsection{ReLU}
Because $\bm{R}$ is a rotation matrix, we get
$$ \parallel \bm{f}_v \cdot \bm{R} \parallel  = \parallel \bm{f}_v \parallel.$$
Thus the revised ReLU operation $\text{ReLU}(\bm{f}_v) = \frac{\parallel \bm{f}_v \parallel}{\text{max}\{ \parallel \bm{f}_v \parallel, C \}}\cdot \bm{f}_v$ is rotation-equivariant. The proof is given below.
$$\begin{aligned}
\text{ReLU}(\bm{f}_v \cdot \bm{R}) &= \frac{\parallel  \bm{f}_v \cdot  \bm{R}\parallel}{\text{max} \{ \parallel \bm{f}_v \cdot  \bm{R} \parallel, C \}} \cdot \bm{f}_v \cdot  \bm{R} \\
&= \frac{\parallel \bm{f}_v \parallel}{\text{max}\{ \parallel \bm{f}_v \parallel, C \}}\cdot \bm{f}_v \cdot  \bm{R}  \\
&= \bm{f}_v \cdot \frac{\parallel \bm{f}_v \parallel}{\text{max}\{ \parallel \bm{f}_v \parallel, C \}}\cdot \bm{R} \\
&= \text{ReLU}(\bm{f}_v) \cdot \bm{R}
\end{aligned}$$

\subsubsection{Batch-normalization}
The revised Batch-normalization $\text{norm}(\bm{f}_v^{(k)}) = \dfrac {\bm{f}_v^{(k)}}{\sqrt{E_{k'}[\parallel \bm{f}_v^{(k')} \parallel ^2]+\epsilon }}$ is rotation-equivariant. The proof is given below.
$$\begin{aligned}
\text{norm}(\bm{f}_v^{(k)} \cdot \bm{R} ) &= \dfrac {\bm{f}_v^{(k)} \cdot \bm{R} }{\sqrt{E_{k'}[\parallel \bm{f}_v^{(k')} \cdot \bm{R}  \parallel ^2]+\epsilon }}\\
	&=  \dfrac {\bm{f}_v^{(k)} \cdot \bm{R}}{\sqrt{E_{k'}[\parallel \bm{f}_v^{(k')} \parallel ^2]+\epsilon }}\\
	&=  \dfrac {\bm{f}_v^{(k)}}{\sqrt{E_{k'}[\parallel \bm{f}_v^{(k')} \parallel ^2]+\epsilon }} \cdot \bm{R}\\
	&=\text{norm}(\bm{f}_v^{(k)}) \cdot \bm{R}
\end{aligned}$$, where $\bm{f}_v^{(k)}$ denotes the feature of the $k$-th sample.

\subsubsection{Avg/Max-pooling/Dropout}
The Avg/Max-pooling operation and dropout layer can be represented as $pool/dropout(\bm{f}) = \bm{f}\circ \bm{m}$, where $\bm{f} \in \mathbb{H}_{d}^{n}$, and $m\in \{0,1\}^{n}$ represents the selection of elements. Hence
$$ \begin{aligned}
\text{pool/dropout}(\bm{f} \cdot \bm{R}) &= (\bm{f} \cdot \bm{R}) \circ m= (\bm{f}\circ m) \cdot \bm{R}\\
&= \text{pool/dropout}(\bm{f})  \cdot \bm{R}
\end{aligned}$$

\subsubsection{Skip connection}
Skip connection can be formulated as $\bm{f}+\Phi(\bm{f})$. If $\Phi(\bm{f})$ is rotation-equivariant then
$$\begin{aligned}
	\bm{f} \cdot \bm{R} + \Phi(\bm{f} \cdot \bm{R})& = \bm{f} \cdot \bm{R} +  \Phi(\bm{f}) \cdot \bm{R}  \\
	&=  (\bm{f} + \Phi(\bm{f})) \cdot \bm{R}
\end{aligned}$$

\section{Experiments: visualization}
Fig.~\ref{fig:cifar} and Fig.~\ref{fig:celeba} in the supplementary material show more visualization results of the Fig. 3 in the paper on CIFAR-10 and CelebA, respectively.

Fig.~\ref{fig:phases} in the supplementary material shows more reconstructed results of the Fig. 4 in the paper when the inversion attacker uses different phases to decrypt the $3$-ary feature. The first image of every two rows is the original input image. Our method performed the best in the privacy protection.
\begin{figure}[!ht]
  \centering
  \includegraphics[width=0.95\linewidth]{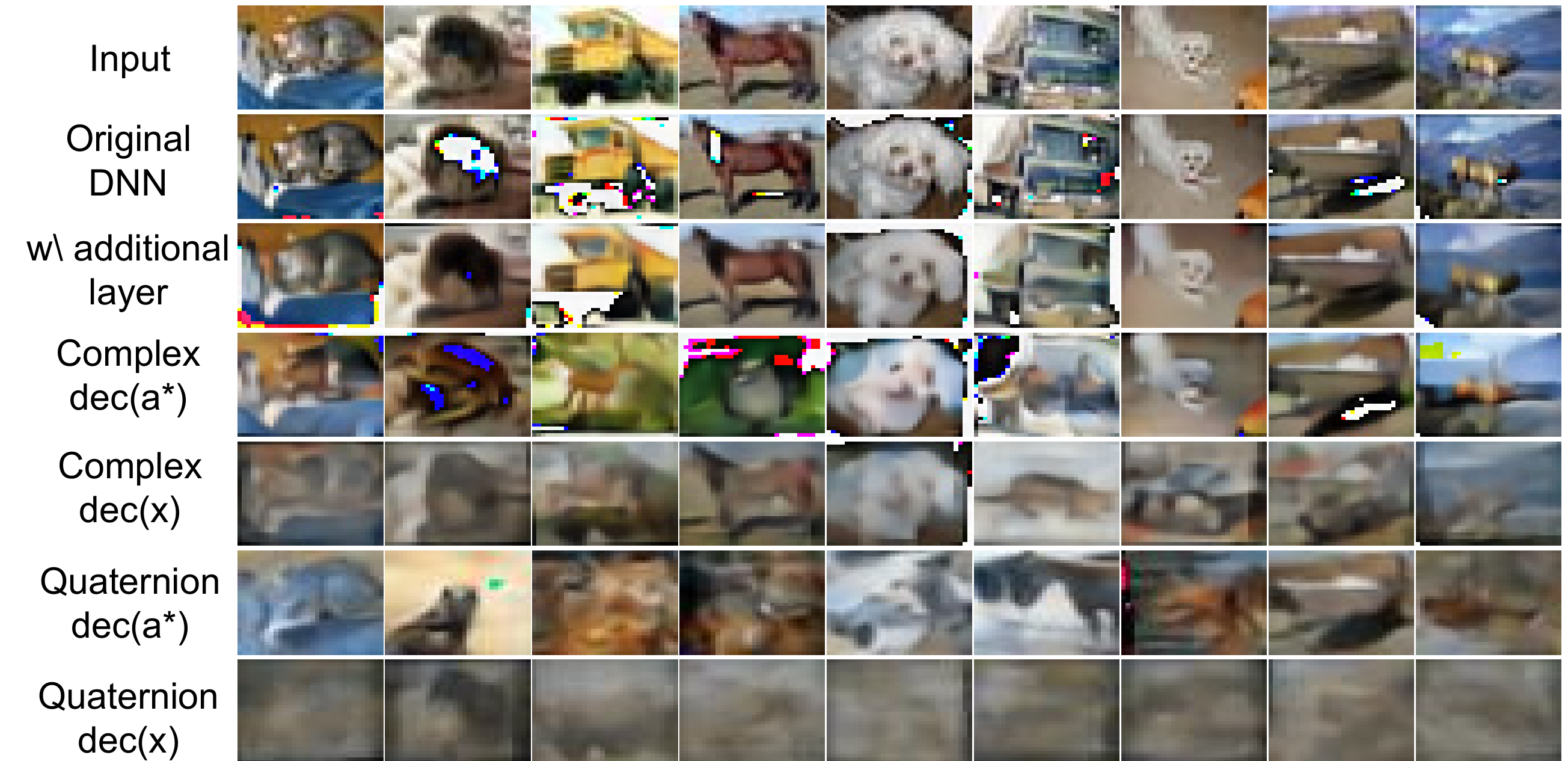}
  \caption{CIFAR-10 images reconstructed from different neural networks.}
  \label{fig:cifar}\vspace{-10pt}
\end{figure}

\begin{figure}[!ht]
  \centering
  \includegraphics[width=\linewidth]{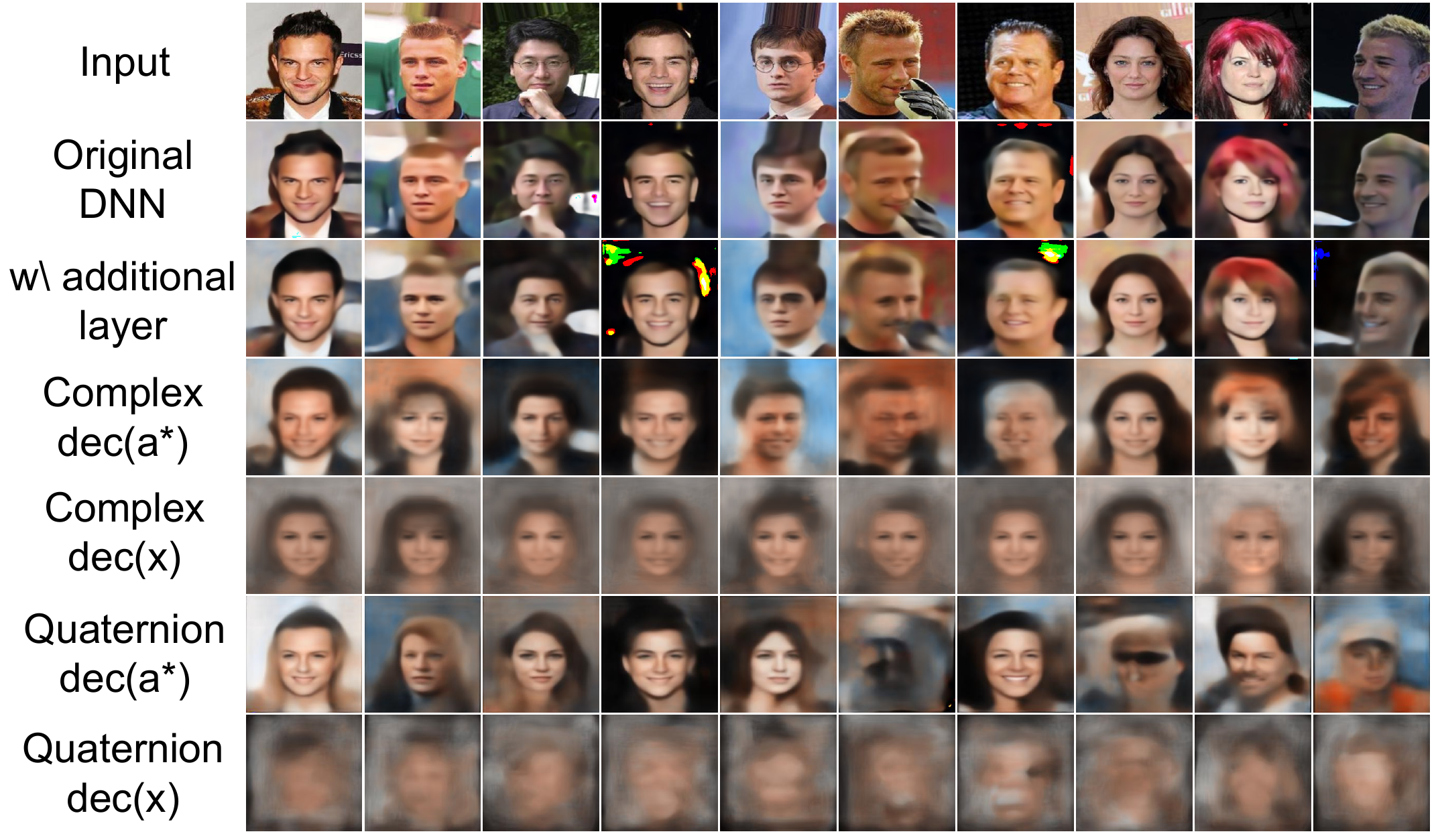}
  \vspace{-5pt}
  \caption{CelebA images reconstructed from different neural networks.}
  \label{fig:celeba}
\end{figure}

\begin{figure}[t]
  \centering
  \includegraphics[width=0.75\linewidth]{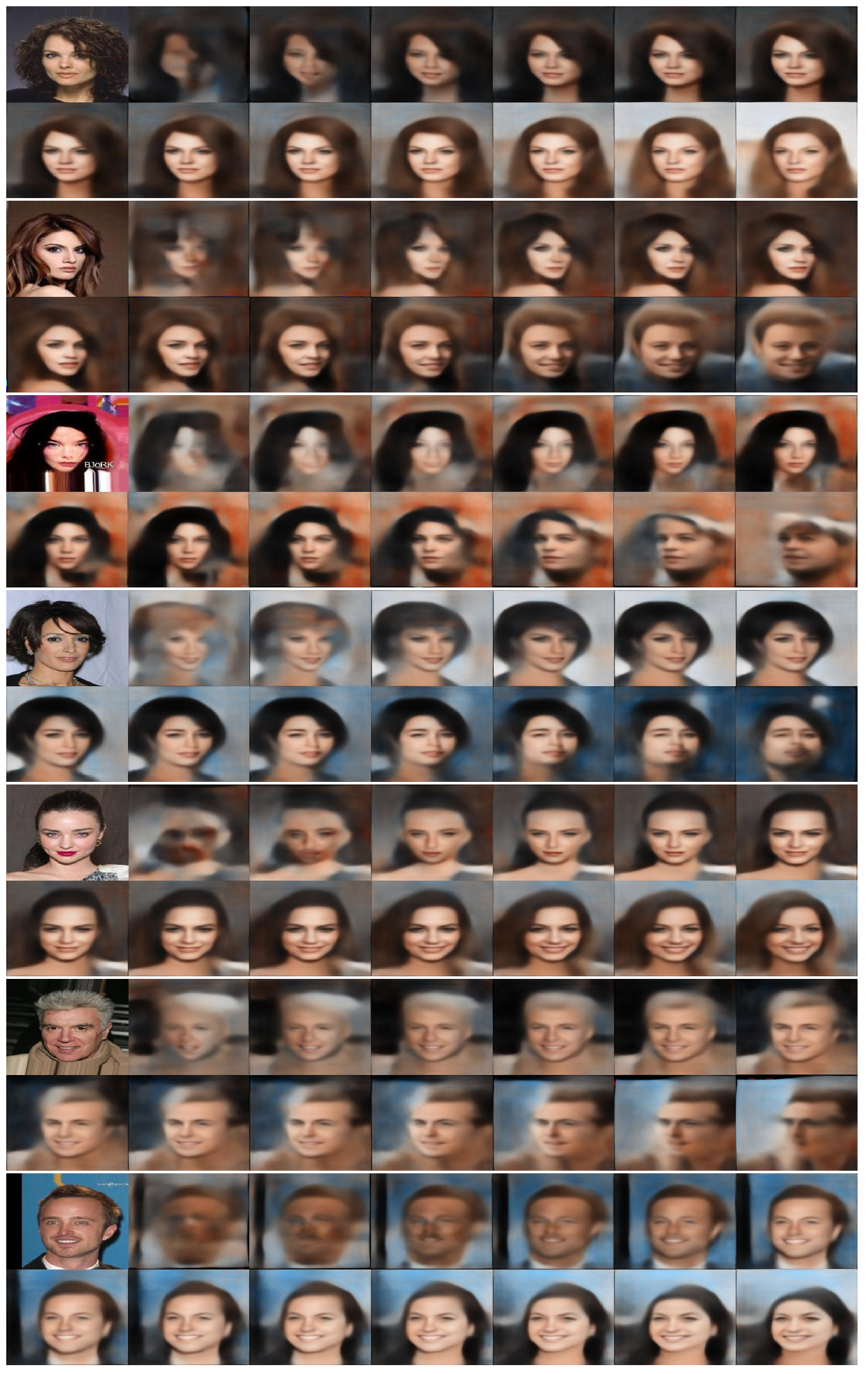}
  \caption{CelebA images reconstructed from different phases. The first image of every two rows is the original input image.}
  \label{fig:phases}
\end{figure}

\clearpage
\section{Computation of the rank of the estimated sample.}
The result of the rank of the estimated samples was shown in Table 3 of the paper. In this section, we introduce the computation of the rank of the estimated samples in different RENNs.

The rank of the estimated sample reflects the number of samples that are more similar to the input than the estimated sample. The estimated sample is generated by inversion attackers, who aim to use different phases to decrypt the encrypted $d$-ary feature to get the input information. $\Delta \theta$ denotes the angle between the phase estimated by attackers and the phase that contains the input information. If the angle between the phase of another sample and the phase contains the input information is less than $\Delta \theta$, then we consider this sample to be more similar to the input than the estimated sample.

\begin{figure}[t]
  \centering
  \includegraphics[width=0.99\linewidth]{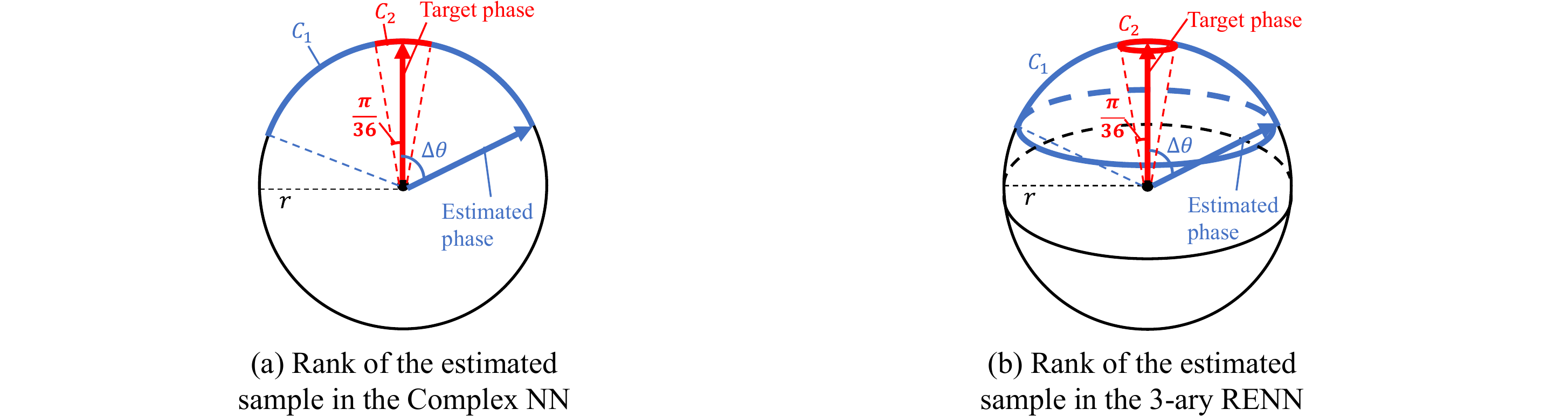}
  \caption{Rank of the estimated sample in (a) the Complex NN, and (b) the 3-ary RENN. The target phase is the phase that contains the input information. $\Delta \theta$ is the angle between the target phase and the estimated phase used by attackers. (a) $C_1$ represents the circular arc, which is determined by $\Delta\theta$. $C_1 = r \cdot \Delta \theta$ contains all phases better than the estimated phase. $C_2 = r \cdot \frac{\pi}{36}$ represents a small circular arc determined by $\frac{\pi}{36}$, phases in which can be considered to represent the same result as the target phase. The ratio $C_1/C_2$ between the length of $C_1$ and the length of $C_2$ reflects the rank of the estimated phase in the Complex NN. (b) $C_1= r \cdot \Delta \theta$ represents the spherical crown that contains all phases better than the estimated phase. $C_2 = r \cdot \frac{\pi}{36}$ represents a small spherical crown, phases in which can be considered to represent the same result. The ratio $C_1/C_2$ between the area of $C_1$ and the area of $C_2$ reflects the rank of the estimated phase.}
  \label{fig:rank}
\end{figure}

We propose the following method to compute the rank of the estimated sample.
For the $d$-ary feature, we can use a point in $d$-dimensional space to represent a phase. As Fig.~\ref{fig:rank} shows, all possible phases of the $d$-ary feature form a $d$-dimensional sphere with a radius $r$. Note that the radius $r$ will be eliminated to compute the rank, so we do not need to consider the value of $r$. In this way, all samples that are more similar to the estimated sample form a spherical crown, whose area is denoted by $C_1$. $\text{Area}(C_1)$ represents the area of the spherical crown $C_1$. According to our experience, if the angle between two phases was less than $\pi/36$, then the corresponding two samples usually represent the same inversion result. Samples that represent the same inversion result can form another spherical crown, whose area is given as $C_2$.
In this way, the number of samples that are more similar to the input than the estimated sample is positively related to the number of $C_2$ contained by $C_1$.
\emph{I.e.} the rank of the estimated sample can be computed as $C_1/C_2$.
Therefore, the rank of the estimated feature in the 3-ary RENN can be computed as follows.
\begin{equation*}
\begin{aligned}
  \text{Rank}_{\text{$3$-ary RENN}}&=\frac{C_1}{C_2} =\frac{\int_{0}^{\Delta\theta} (2{\pi} r \sin\theta)r \,d\theta}{\int_{0}^{\frac{\pi}{36}} (2{\pi} r\sin\theta)r \,d\theta}\\
 & =\frac{2\pi r^2(1-\text{cos}\Delta\theta)}{2\pi r^2(1-\text{cos}\frac{\pi}{36})} \\
 &=\frac{1-\text{cos}\Delta\theta}{1-\text{cos}(\frac{\pi}{36})}
\end{aligned}
\end{equation*}
As for the 5-ary RENN, computing the area of a 5-dimensional spherical crown needs the area of a 4-dimensional sphere, which is $2\pi^2r^3$. $r$ is the radius of the sphere. Thus, the rank of the estimated feature in the 5-ary RENN can be computed as follows.
\begin{equation*}
\begin{aligned}
  \text{Rank}_{\text{$5$-ary RENN}}&=\frac{C_1}{C_2} =\frac{\int_{0}^{\Delta\theta} 2{\pi}^{2} (r \sin\theta)^{3} r \,d\theta}{\int_{0}^{\frac{\pi}{36}} 2{\pi}^{2} (r \sin\theta)^{3} r \,d\theta}\\
  & =\frac{(\frac{2}{3}-\cos\Delta\theta+ \frac{1}{3}\cos^{3}\Delta\theta){\pi}^{2} r^4}{(\frac{2}{3}-\cos\frac{\pi}{36}+ \frac{1}{3}\cos^{3}\frac{\pi}{36}){\pi}^{2} r^4} \\
  &=\frac{2-3\cos\Delta\theta+ \cos^{3}\Delta\theta} {2-3\cos\frac{\pi}{36}+ \cos^{3}\frac{\pi}{36}}
\end{aligned}
\end{equation*}
Similarly, for the Complex NN, we can consider it as the 2-ary RENN. As Fig.\ref{fig:rank} (a) shows, the sphere in the 2-dimensional space is a circle, and the spherical crown in the 2-dimensional space is a circular arc. The area of the spherical crown can be computed as the length of the circular arc. Thus, the rank can be computed as follows.
\begin{equation*}
  \text{Rank}_{\text{Complex}}=\frac{C_1}{C_2} =\frac{2\Delta \theta r}{2\times\frac{\pi}{36} r}=\frac{36\Delta\theta}{\pi}
\end{equation*}
We compute the rank for each estimated sample, and report the average rank over all samples in Table 3 of the paper.

\section{Special cases of RENNs}

\subsection{Complex-valued neural networks (Complex-valued NNs)}

Let $(I, y)\in \mathcal{D}$ denote an input and its label in the training dataset, and let $g$ be the encoder at the local device. Given the input $I$, the intermediate-layer feature is computed as
\begin{equation}
a = g(I) \in \mathbb{R}^n,
\end{equation}
but we do not directly submit $a$ to the processing module. Instead, we introduce a fooling counterpart $b$ to construct a complex-valued feature as follows:
\begin{equation}
\label{eqn1:encrypt}
x = \exp(i\theta)\big[a+bi\big],
\end{equation}
where $\theta$ and $b\in\mathbb{R}^n$ are randomly chosen. $b$ is the fooling counterpart, which does not contain any private information of $a$, but its magnitude is comparable with $a$ to cause obfuscation. The encoded feature is then sent to the processing module $\boldsymbol\Phi$, which produces the complex-valued feature $h={\boldsymbol\Phi}(x)$. Upon receiving $h$, the decoder makes prediction $\hat{y}$ on $I$ by inverting the complex-valued feature $h$ back:
\begin{equation}
\label{eqn1:rob}
\hat{y} = d(\Re[h \cdot \exp(-i\theta)]),
\end{equation}
where $d$ denotes the decoder module, which can be constructed as either a shallow network or just a softmax layer. $\Re(\cdot)$ denotes the operation of picking real parts of complex values.

The core design of the processing module is to allow the complex-valued feature $h={\boldsymbol\Phi}(x)$ to be successfully decoded later by the decoder. \emph{I.e.} if we rotate the complex-valued feature $a + bi$ by an angle $\theta$, all the features of the following layers are supposed to be rotated by the same angle. We represent the processing module as the cascaded functions of multiple layers ${\boldsymbol\Phi}(x)=\Phi_{n}(\Phi_{n-1}(\cdots\Phi_1(x)))$, where $\Phi_{j}(\cdot)$ denotes the function of the $j$-th layer; $f_{j}=\Phi_{j}(f_{j-1})$ represents the output of the $j$-th layer. Thus the processing module should have the following property:
\begin{equation}
	\label{eqn1:rotate}
	\boldsymbol{\Phi}(f^{(\theta)}) = e^{i\theta} \boldsymbol{\Phi}(f) \quad \text{s.t.}\quad f^{(\theta)} \triangleq e^{i\theta}f,~\forall \theta \in [0, 2\pi).
\end{equation}
In other words, the function of each intermediate layer in the processing module should satisfy
\begin{equation}
	\label{eqn1:layer_rotate}
	\Phi_{j}(f_{j-1}^{(\theta)}) = e^{i\theta} \Phi_{j}(f_{j-1}) \\
	\quad \text{s.t.}\quad f_{j-1}^{(\theta)} \triangleq  e^{i\theta}f_{j-1},\;\forall j \in \{2, \ldots, n\}, ~\forall \theta \in [0, 2\pi).
\end{equation}
to recursively prove Eqn.~(\ref{eqn1:rotate}).

\subsection{Quaternion-valued neural networks (QNNs)}

\textbf{Quaternion:} Quaternion is a number system extended from the complex number. Unlike the complex number, a quaternion consists of three imaginary parts $q_1\bm{i}$, $q_2\bm{j}$, $q_3\bm{k}$, and one real part $q_0$, which is given as $\bm{q}=q_0+q_1 \bm{i}+q_2\bm{j}+q_3\bm{k}$. If the real part of a quaternion is zero ($q_0=0$), we call it a pure quaternion. The quaternion subject to $||\bm{q}|| = \sqrt{q_0^2+q_1^2+q_2^2+q_3^2}=1$ is termed a unit quaternion.
The products of basis elements $\bm{i}$, $\bm{j}$, $\bm{k}$ are given as $\bm{i}^2 = \bm{j}^2 = \bm{k}^2 = \bm{ijk} = -1$, and $\bm{ij}=\bm{k}$,  $\bm{jk}=\bm{i}$,  $\bm{ki}=\bm{j}$, $\bm{ji}=\bm{-k}$, $\bm{kj}=\bm{-i}$, $\bm{ik}=\bm{-j}$.
Note that the multiplication of two imaginary parts is non-commutative, \emph{i.e.} $\bm{ij} \neq \bm{ji}$, $\bm{jk} \neq \bm{kj}$, $\bm{ki} \neq \bm{ik}$. Each quaternion has a polar decomposition. The polar decomposition of a unit quaternion is defined as $\emph{e}^{\bm{o}\frac{\theta}{2}}=\text{cos}\frac{\theta}{2}+\text{sin}\frac{\theta}{2}(o_1\bm{i}+o_2\bm{j}+o_3\bm{k})$, \emph{s.t.} $\sqrt{o_1^2+o_2^2+o_3^2}=1$.

When we use a pure quaternion $\bm{q} = 0+ q_1\bm{i}+ q_2\bm{j}+ q_3\bm{k}$ to represent a point $[q_1,q_2,q_3]^T \in \mathbb{R}^3 $ in a 3D space, the rotation of the point around the axis $\bm{o}=o_1\bm{i}+o_2\bm{j}+o_3\bm{k}$, \emph{s.t.} $\sqrt{o_1^2+o_2^2+o_3^2}=1$, by the angle $\theta$ can be represented as $\bm{Rq\overline{R}}$, where $\bm{R}= \textit{e}^{\bm{o}\frac{\theta}{2}}$, and $\bm{\overline{R}}=\textit{e}^{-\bm{o}\frac{\theta}{2}}$ is the conjugation of $\bm{R}$.

Given a pure quaternion-valued vector $\bm{x}=0+a\bm{i}+b\bm{j}+c\bm{k}\in\mathbb{H}^{n}$, and a real-valued vector $w\in\mathbb{R}^n$, we have

\vspace{-15pt}
\begin{small}
\begin{equation}
\label{eqn2:multi}
\bm{x}^Tw=\sum_{v=1}^n \bm{x}_vw_v=0+(a^Tw)\bm{i}+(b^Tw)\bm{j}+(c^Tw)\bm{k}
\end{equation}
\end{small}
\vspace{-15pt}

\textbf{Design of the QNN:} We introduce a set of basic rules to transform a traditional neural network into a QNN. We only revise the traditional real-valued feature to the quaternion-valued feature. In comparison, parameters in the QNN, \emph{e.g.} weights in a filter, are still real numbers, instead of quaternions.

\textit{Encoder:} Given an input $I\in \textbf{I}$, the encoder module $g$ computes a traditional real-valued feature $a$, as follows.
\begin{small}
\begin{equation}
\centering
\label{eqn2:encoder}
a=g(I) \in \mathbb{R}^n
\end{equation}
\end{small}
Then the encoder module uses $a$ and two fooling counterparts $b$, $c$ to generate a quaternion-valued feature $\bm{x} = 0+a\bm{i}+b\bm{j}+c\bm{k}\in \mathbb{H}^n$. Each element in $\bm{x}$ is a quaternion. Note that we can equivalently let $b=g(I)$ or $c=g(I)$ without loss of generality. We encrypt the quaternion-valued feature by rotating $\bm{x}$ along a random axis $\bm{o}=0+o_1\bm{i}+o_2\bm{j}+o_3\bm{k}$ by a random rotation angle $\theta$,
$o_1, o_2, o_3 \in \mathbb{R}$, $||\bm{o}||=1$, and obtain the encrypted feature $\bm{f}\in\mathbb{H}^n$, as follows.
\begin{small}
\begin{equation}
\centering
\label{eqn2:encrypt}
\bm{f}=\Psi_{\bm{R}}(a)=\bm{R}\circ\bm{x}\circ\bm{\overline{R}}=\bm{R}\circ(0+a\bm{i}+b\bm{j}+c\bm{k})\circ\bm{\overline{R}}
\end{equation}
\end{small}
where $\Psi_{\bm{R}}(\cdot)$ denotes the function which applies a random rotation $\bm{R}$ to the original quaternion-valued feature $\bm{x}$, $\bm{R}=e^{\bm{o}\frac{\theta}{2}}=\text{cos}\frac{\theta}{2}+\text{sin}\frac{\theta}{2}(o_1\bm{i}+o_2\bm{j}+o_3\bm{k})$, and
$\circ$ denotes the element-wise multipication. The encrypted feature $\bm{f}$ will be sent to the processing module $\Phi$. In this way, we can consider $\bm{Ri}\overline{\bm{R}}$ as the target phase, which encodes the input information, and $\bm{R}$ can be taken as the private key.

\textit{Processing module:} Inspired by homomorphic encryption, we revise the operation of each layer in the processing module to satisfy rotation equivariance of the quaternion-valued feature. The rotation equivariance property ensures that the input information is always encoded in the same phase of all quaternion-valued features of all layers in the processing module. In this way, the decoder module can use the target phase to decrypt the input information from the quaternion-valued feature.

The rotation equivariance property can be summarized, as follows. If we use $\bm{R}\circ\bm{x}\circ\bm{\overline{R}}$ to rotate quaternion-valued feature $\bm{x}$ along the axis $\bm{o}$ by the angle $\theta$, then quaternion-valued feature elements in each intermediate layer of the processing module are supposed to be rotated along the same axis by the same angle, as follows.
\begin{small}
\begin{equation}
\label{eqn2:module_rotate}
\bm{\Phi}(\bm{R}\circ\bm{x}\circ\bm{\overline{R}})=\bm{R}\circ\bm{\Phi}(\bm{x})\circ\bm{\overline{R}}
\end{equation}
\end{small}
Let us consider $\bm{h}_0=\bm{\Phi}(\bm{x})$ without the rotation as the output of the processing module, \emph{i.e.} $\theta=0$, $\bm{R}=e^{\bm{o}\frac{\theta}{2}}=\bm{1}$. The input information is hidden in the imaginary part $\bm{i}$.
Since all parameters in the processing module are real-valued, according to Eqn.~\eqref{eqn2:multi}, the output of the processing module can be represented in the form
\begin{small}
\begin{equation}
\label{eqn2:processing1}
\bm{h}_0=\bm{\Phi}(\bm{x})=0+(Aa)\bm{i}+(Ab)\bm{j}+(Ac)\bm{k}\text{.}
\end{equation}
\end{small}
$A$ is a real-valued matrix that represents effects that combine all non-linear transformations in $\bm{\Phi}(\bm{x})$, when $\bm{\Phi}(\bm{x})$ only uses ReLU as non-linear layers. Please see the supplementary material for the computation of $A$. In this way, the input information is still hidden in the imaginary part $\bm{i}$ of $\bm{h}_0$.  Then, let us consider the rotation $\bm{R}$, $\bm{h}=\bm{\Phi}(\bm{R}\circ\bm{x}\circ\overline{\bm{R}})$.
According to Eqn.~\eqref{eqn2:module_rotate} and Eqn.~\eqref{eqn2:processing1}, the output is given as
\begin{small}
\begin{equation}
  \begin{aligned}
  \label{eqn2:processing2}
  \bm{h}&=\bm{\Phi}(\bm{R}\circ\bm{x}\circ\overline{\bm{R}})=\bm{R}\circ\bm{h}_0\circ\overline{\bm{R}}\\
  &=(\bm{Ri}\overline{\bm{R}})\circ(Aa)+(\bm{Rj}\overline{\bm{R}})\circ(Ab)+(\bm{Rk}\overline{\bm{R}})\circ(Ac)
\end{aligned}
\end{equation}
\end{small}
In this way, the input information is hidden in the phase $\bm{Ri}\overline{\bm{R}}$ of $\bm{h}$. To ensure the above rotation equivariance, we recursively ensure rotation equivariance of the layerwise operation of each layer inside the processing module. The processing module can be represented as cascaded layers $\Phi(\bm{f})=\Phi_L(\Phi_{L-1}(\cdots\Phi_1(\bm{f}))$, where $\Phi_l(\cdot)$ denotes the $l$-th layer in the processing model. Let $\bm{f}'$ denote the input feature of the $l$-th layer, then the layerwise operation is supposed to satisfy
\begin{small}
\begin{equation}
\label{eqn2:single_rotate}
\Phi_{l}(\bm{R}\circ\bm{f}'\circ\bm{\overline{R}})=\bm{R}\circ\Phi_{l}(\bm{f}')\circ\bm{\overline{R}}\text{.}
\end{equation}
\end{small}
Thus, this equation recursively ensures rotation equivariance in Eqn.~\eqref{eqn2:module_rotate}.

\textit{Decoder:} Let $\bm{h}=\bm{\Phi}(\bm{f})$. Let $d$ denote the decoder module, which can be implemented as a shallow network or a simple softmax layer. The decoder module can get the final result $\hat{y}$ as follows.
\begin{small}
\begin{equation}
\label{eqn2:decoder}
\hat{y}=d(\Psi_{\bm{R}}^{-1}(\bm{h})),\quad\Psi_{\bm{R}}^{-1}(\bm{h})=\text{Im}_{\bm i}(\overline{\bm{R}}\circ\bm{h}\circ\bm{R})
\end{equation}
\end{small}
where $\Psi_{\bm{R}}^{-1}(\cdot)$ indicates the inverse function of $\Psi_{\bm{R}}(\cdot)$. The rotation in Eqn.~\eqref{eqn2:decoder} is the inverse of the rotation in Eqn.~\eqref{eqn2:encrypt}. $\text{Im}_{\bm{i}}(\cdot)$ denotes the operation that picks the $\bm{i}$ part from quaternions, and returns a real-valued feature.

\subsection{Complex-valued NN and QNN are special cases of RENNs}

\textit{For the complex-valued NN:} We can rewrite the complex-valued feature $a+b\bm{i}$ as a 2-ary feature $\bm{f}\in\mathbb{H}_2^n$, in which $a,b\in\mathbb{R}^n$ are taken as the two components. Accordingly, the rotation of the complex-valued feature $a+b\bm{i}$ by $\exp[\theta\bm{i}]$ can be represented in the scenario of 2-ary features as follows.
\begin{equation}
\exp[\theta\bm{i}]\cdot\big[a+b\bm{i}\big]\quad\textrm{corresponds to}\quad \bm{R}\circ\bm{f},\quad\textrm{where}\quad\bm{R}=\left[\begin{array}{cc}\cos(\theta)&-\sin(\theta)\\\cos(\theta)&\sin(\theta)\\\end{array}\right].
\end{equation}
It is because
\begin{equation}
\exp[\theta\bm{i}]\cdot\big[a+b\bm{i}\big]=[a\cdot\cos(\theta)-b\cdot\sin(\theta)]+\bm{i}[a\cdot\sin(\theta)+b\cdot\cos(\theta)]\\
\end{equation}

\textit{For the QNN:} We can rewrite the quaternion-valued feature $a\bm{i}+b\bm{j}+c\bm{k}$ as a 3-ary feature $\bm{f}\in\mathbb{H}_3^n$, in which $a,b,c\in\mathbb{R}^n$ are taken as the three components. Accordingly, the rotation of the quaternion-valued feature $a\bm{i}+b\bm{j}+c\bm{k}$ by $e^{\bm{o}\frac{\theta}{2}}=\text{cos}\frac{\theta}{2}+\text{sin}\frac{\theta}{2}(o_1\bm{i}+o_2\bm{j}+o_3\bm{k})$ can be represented in the scenario of 3-ary features as follows.
\begin{equation}
\exp[{\bm{o}\frac{\theta}{2}}]
\circ\big[a\bm{i}+b\bm{j}+c\bm{k}\big]\circ\exp[-{\bm{o}\frac{\theta}{2}}]\quad\textrm{corresponds to}\quad \bm{R}\circ\bm{f},
\end{equation}
where,
\begin{small}
\begin{equation}
\bm{R}\!\!=\!\!\left[\!\!\!\!\begin{array}{ccc}
-\sin^2(\frac{\theta}{2})[o_2^2+o_3^2]
\!\!&\!\! -\cos(\frac{\theta}{2})\sin(\frac{\theta}{2})o_3+\sin^2(\frac{\theta}{2})o_1 o_2
\!\!&\!\! \cos(\frac{\theta}{2})\sin(\frac{\theta}{2})o_2+\sin^2(\frac{\theta}{2})o_1 o_3
\\
\cos(\frac{\theta}{2})\sin(\frac{\theta}{2})o_3+\sin^2(\frac{\theta}{2})o_1 o_2
\!\!&\!\! -\sin^2(\frac{\theta}{2})[o_1^2+o_3^2]
\!\!&\!\! -\cos(\frac{\theta}{2})\sin(\frac{\theta}{2})o_1+\sin^2(\frac{\theta}{2})o_2 o_3
\\
-\cos(\frac{\theta}{2})\sin(\frac{\theta}{2})o_2+\sin^2(\frac{\theta}{2})o_1 o_3
\!\!&\!\! \cos(\frac{\theta}{2})\sin(\frac{\theta}{2})o_1+\sin^2(\frac{\theta}{2})o_2 o_3
\!\!&\!\! -\sin^2(\frac{\theta}{2})[o_1^2+o_2^2]
\\
\end{array}\!\!\!\!\right].
\end{equation}
\end{small}

\subsection{Transferring parameters from the complex-valued NN to RENNs ($d\geq2$)}

$\bullet$ In previous subsections, we have proved that the complex-valued NN and the QNN can be represented as special cases of RENNs when $d=2$ and $d=3$, respectively. Therefore, it is easy to know that \textbf{we can directly transfer parameters in a well-trained complex-valued NN into a RENN with $d=2$}.

$\bullet$ Instead of showing how to define the rotation matrix $R$ to enable a RENN with $d>2$ to use parameters of a complex-valued NN (a RENN with $d=2$), in this subsection, let us focus on a more generic problem, \emph{i.e.} \textbf{how to define the rotation matrix $R$ to enable a $d$-ary RENN to use parameters of a $d'$-ary RENN when $d>d'$}.

Let us consider the inference process of the $d'$-ary RENN. Let $\bm{f}'=\bm{R}'\circ\bm{x}'\in\mathbb{H}_{d'}^n$ denote the encrypted feature of the $d'$-ary RENN, where $\bm{R}'\in\mathbb{R}^{d'\times d'}$ is referred to as the rotation matrix. Let $\bm{f}'_v$ denote a specific element in the $d'$-ary feature $\bm{f}$, which can also be represented as a $d'$-dimensional vector. Thus, we can use the matrix $F'\in\mathbb{R}^{d'\times n}$ to represent the $d'$-ary feature $\bm{f}'$. Accordingly, the matrix $X'\in\mathbb{R}^{d'\times n}$ corresponds to the $d'$-ary feature $\bm{x}'$. In this way, the convolution operation (without the bias term) of $\bm{f}'$ can be written as $\Phi(F')=F'\cdot W\in\mathbb{R}^{d'\times m}$, where $W'\in\mathbb{R}^{n\times m}$. \emph{I.e.} we get
\begin{equation}
F'_{d'\times n}=\bm{R}'_{d'\times d'}X_{d'\times n},\qquad \Phi(F')_{d'\times m}=F'_{d'\times n}W_{n\times m}.
\end{equation}

Then, let us consider how to run a $d$-ary RENN with parameters in the $d'$-ary RENN ($d>d'$). In this case, the input feature $\bm{x}$ contains $d-d'$ additional components. We set these $d-d'$ components as $\bm{0}$, and set the rotation matrix $\bm{R}$, as follows.
\begin{equation}
\bm{R}=\left[\begin{array}{cc}
\bm{R}'_{d'\times d'} & \bm{0}_{d\times(d-d')}\\
\bm{0}_{(d-d')\times d} & \bm{0}_{(d-d')\times(d-d')}\\
\end{array}
\right]_{d\times d},\qquad
X=\left[{X'_{d'\times n}\atop \bm{0}_{(d-d')\times n}}\right]_{d\times n}
\end{equation}
In this case, we have
\begin{small}
\begin{equation}
\begin{split}
&F=\bm{R}X=\left[\begin{array}{cc}
\bm{R}'_{d'\times d'} & \bm{0}_{d\times(d-d')}\\
\bm{0}_{(d-d')\times d} & \bm{0}_{(d-d')\times(d-d')}\\
\end{array}
\right]_{d\times d}\cdot\left[{X'_{d'\times n}\atop \bm{0}_{(d-d')\times n}}\right]_{d\times n}
=\left[{\bm{R}'_{d'\times d'}X'_{d'\times n}\atop \bm{0}_{(d-d')\times n}}\right]_{d\times n}=\left[F'_{d'\times n}\atop \bm{0}_{(d-d')\times n}\right]_{d\times n}
\\
&\Phi(F)=F_{d\times n}W_{n\times m}=\left[F'_{d'\times n}\atop \bm{0}_{(d-d')\times n}\right]W_{n\times m}=\left[F'_{d'\times n}W_{n\times m}\atop \bm{0}_{(d-d')\times n}\right]=\left[\Phi(F')_{d'\times n}\atop \bm{0}_{(d-d')\times n}\right]_{d\times n}.
\end{split}
\end{equation}
\end{small}
Thus, given the specific rotation matrix $\bm{R}$, the signal processing in the $d$-ary RENN is the same as that in the $d'$-ary RENN, which proves that we can use the $d$-ary RENN make inference.

\subsection{The value of $d$ vs. the capacity of privacy protection}

When we use a RENN with more components (\emph{i.e.} setting a large value of $d$), the RENN contains more fooling counterparts, which leads to a higher capacity of privacy protection. However, meanwhile, the more components also boost the difficulty of learning a RENN, because in this case, network parameters need to simultaneously deal with more noisy data.
\end{document}